\pdfoutput=1
\documentclass[10pt,twocolumn,letterpaper]{article}

\usepackage[pagenumbers]{wacv}

\usepackage{amsmath}
\usepackage{amssymb}
\usepackage{multirow}
\usepackage{gensymb}
\usepackage{dblfloatfix}

\usepackage{algorithm}
\usepackage{algpseudocode}

\definecolor{wacvblue}{rgb}{0.21,0.49,0.74}
\usepackage[pagebackref,breaklinks,colorlinks,allcolors=wacvblue]{hyperref}

\def\wacvPaperID{1380}
\def\confName{WACV}
\def\confYear{2027}

\title{You Only Flow Once: Calibrated and Real-Time Radar Pose Estimation with Multi-Hypothesis Normalizing Flows}

\author{
Jonas Leo Mueller\textsuperscript{1,3,5} \hspace{0.4em}
Sebastian Hoefler\textsuperscript{3,5} \hspace{0.4em}
Dario Zanca\textsuperscript{3} \hspace{0.4em}
Naga Venkata Sai Jitin Jami\textsuperscript{3,5} \\[2pt]
Thomas Altstidl\textsuperscript{2,3,5} \hspace{0.4em}
Bjoern M. Eskofier\textsuperscript{1,2,3,4,5}
\\[0.3em]
{\footnotesize \textsuperscript{1}Institute for AI in Medicine, LMU Hospital, Munich, Germany \enspace \textsuperscript{5}Munich Center for Machine Learning (MCML), Munich, Germany}\\
{\footnotesize \textsuperscript{2}Chair of AI-supported Therapy Decisions, Ludwig-Maximilians-Universit\"at M\"unchen, Munich, Germany}\\
{\footnotesize \textsuperscript{3}Department Artificial Intelligence in Biomedical Engineering, Friedrich-Alexander-Universit\"at Erlangen-N\"urnberg (FAU), Erlangen, Germany}\\
{\footnotesize \textsuperscript{4}Institute of AI for Health, Helmholtz Zentrum M\"unchen, Neuherberg, Germany \enspace \href{mailto:jonas.leo.mueller@fau.de}{jonas.leo.mueller@fau.de}}
}

\AddToShipoutPicture*{\AtPageUpperLeft{\put(\LenToUnit{0.75in},\LenToUnit{-0.62in}){\parbox[t]{7in}{\centering\fontsize{6.5}{7.5}\selectfont
Accepted version of a paper to appear in the 2027 IEEE/CVF Winter Conference on Applications of Computer Vision (WACV). \textcopyright~2027 IEEE. Personal use of this material is permitted. Permission from IEEE must be obtained for all other uses, in any current or future media, including reprinting/republishing this material for advertising or promotional purposes, creating new collective works, for resale or redistribution to servers or lists, or reuse of any copyrighted component of this work in other works.}}}}

\begin{document}
\maketitle

\begin{abstract}
Sparse and noisy millimeter-wave radar point cloud observations often correspond to multiple plausible human poses, making deterministic pose estimation fundamentally ill-posed.
Yet existing radar methods remain deterministic, collapsing this ambiguity into a single estimate. Diffusion-based alternatives can model multi-hypothesis distributions but require costly sequential denoising for each sample and lack calibrated uncertainty.
We propose Multi-Hypothesis Normalizing Flow Pose Generator (MH-NFPG), which models pose distributions from radar point clouds using a conditional normalizing flow.
Specifically, we combine a spatiotemporal transformer backbone with a normalizing flow that transforms a Laplace base distribution into an expressive posterior, generated in parallel through a single forward pass.
Leveraging this efficiency, we outperform diffusion-based alternatives in calibration across three radar benchmarks (MM-Fi, mmRadPose, mRI) at comparable pose accuracy, while achieving up to 22$\times$ faster inference and reducing calibration error by up to 84\%.
We find that calibration degrades substantially for diffusion models, whereas our flow-based approach maintains reliable coverage, also across environments.
These results demonstrate normalizing flows as a practical alternative to diffusion models for real-time, uncertainty-aware radar pose estimation. Our code is available at {\def\UrlBreaks{\do\/}\url{https://github.com/jonasmueler/YouOnlyFlowOnce_WACV}}.
\end{abstract}

\section{Introduction}
\label{sec:intro}

\begin{figure}[t!]
    \centering
    \includegraphics[width=\linewidth, trim={0 0 0 22}, clip]{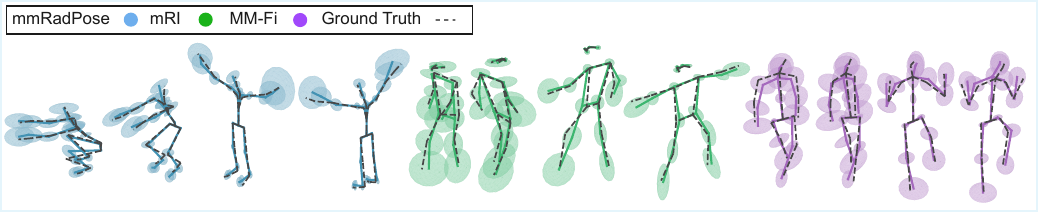}
    \caption{\textbf{MH-NFPG} produces calibrated 3D pose distributions from radar point clouds in a single forward pass, with predictions (colored), ground truth (dashed), and 50\% confidence ellipsoids on mmRadPose (blue), mRI (green), and MM-Fi (purple).}
    \label{fig:qualitative_actions}
\end{figure}

Uncertainty quantification underpins trustworthy machine learning, from medical decisions~\cite{begoli2019need} to computer vision~\cite{kendall2017uncertainties, abdar2021review}. It is no less vital for radar human pose estimation, a privacy-preserving, lighting-robust modality~\cite{sun2019privacy, ho2024rt, lee2023hupr, fan2024diffusion, yang2024mmfi, chen2022mmbody} whose safety-critical use makes confident errors costly. An overconfident pose can corrupt a clinician's assessment in rehabilitation monitoring~\cite{10.1007/978-3-032-00652-3_17}, or drive a robot into the worker it should avoid in human-robot collaboration~\cite{10758369}. How strongly a body part scatters the incident waves is set by its radar cross section (RCS) $\sigma_{\mathrm{rcs}}$, which is small for distal parts such as the hands and feet, so they produce faint echoes, and the received power falls off with range as $R^{-4}$~\cite{richards2005fundamentals}. Physically, the RCS is the effective reflecting area a body part presents to the radar, set by its size, shape, and orientation relative to the sensor, so the small, curved, fast-moving extremities present the least area and are exactly the joints whose returns are weakest. For real-time use, a Constant False Alarm Rate (CFAR) detector thresholds the dense raw radar tensor into a sparse point cloud~\cite{richards2005fundamentals, 10904474} and can drop these weak returns, while multipath and scattering can further influence the points that remain~\cite{NEURIPS2022_af9c9c6d}.

Recovering 3D pose from such a point cloud is therefore an ill-posed inverse problem that we learn to solve. The forward map from pose to measurement is lossy and many-to-one, so one point cloud matches many plausible poses. The resulting uncertainty is shaped by the sensor physics, and changes with the recording environment, the subject, and the radar aspect angle. A deployable estimator should therefore report not a single guess but a calibrated distribution over the plausible poses, one that stays calibrated even when the test environment was not seen during training.

Despite these sensor characteristics, radar pose estimation from point clouds is still almost entirely deterministic. Standard regressors collapse the ambiguous signal into one compromise estimate~\cite{9083948, 10904474, NEURIPS2022_af9c9c6d, an2022fast, xue2021mmmesh, chen2022mmbody} and discard uncertainty, and the uncertainty quantification work that exists operates on full five-dimensional complex radar tensors rather than the point cloud~\cite{11729073}, a far heavier input to transmit on device, and requires recalibration after training. Diffusion models can in principle represent ambiguity by drawing many hypotheses~\cite{holmquist2023diffpose, gong2023diffpose, feng2023diffpose}, yet producing $N$ hypotheses needs $N \times M$ sequential denoising steps, which is too slow for real-time use. Recent radar diffusion work~\cite{fan2024diffusion} avoids this cost with deterministic inference, but at the cost of stochasticity and uncertainty quantification (Section~\ref{sec:related}).

No prior radar method provides both at once, a calibrated multi-hypothesis posterior over the ambiguous point-cloud-to-pose mapping and real-time inference. Normalizing flows (NFs) can close this gap. Their bijective structure gives an exact likelihood for training, and batching can draw every hypothesis in one forward pass. NFs have been applied to RGB pose estimation~\cite{wehrbein2021probabilistic, han2025propose}, where the NF is conditioned on structured 2D keypoints. Radar offers no comparable keypoint detector, so we instead derive the conditioning signal from a prior learned on the radar itself, namely a heteroscedastic multivariate Gaussian~\cite{11729073}. Our model, the Multi-Hypothesis Normalizing Flow Pose Generator (MH-NFPG), combines this prior with a conditional NF. A permutation-invariant spatiotemporal transformer backbone ingests the unordered, variable-size point set and parameterizes the prior, whose full covariance captures the anisotropic, input-dependent uncertainty of radar detections. A temporal graph-convolutional conditioning branch then propagates prior motion context across frames, so low-radar-cross-section joints such as hands and feet stay constrained on the frames where they go undetected. The conditional Real NVP NF~\cite{dinh2017density} starts from a Laplace base distribution, whose heavier tails than a Gaussian accommodate multipath and ghost-target outliers, and transforms it through affine coupling layers into the full pose distribution.
Our contributions are as follows.
\begin{itemize}
    \item \textbf{Intrinsic calibration.} Exact-likelihood training yields empirically calibrated distributions with no post-hoc recalibration, on all three datasets, with 34\% to 84\% lower expected calibration error than the diffusion baselines, and holds even when an entire environment is held out.
    \item \textbf{Real-time multi-hypothesis inference.} A full posterior is produced in one forward pass at about 93 frames per second, up to 22$\times$ faster than the stochastic diffusion baselines.
    \item \textbf{No accuracy trade-off.} On all three datasets our MPJPE and PA-MPJPE are on par with the strongest baseline.
\end{itemize}
Our code is publicly available on GitHub.

\section{Related Work}
\label{sec:related}

\subsection{Multi-Hypothesis Pose Estimation}
Multi-hypothesis modelling (MHM) was introduced for inherently ambiguous prediction tasks, where one input admits several valid outputs, by predicting a set of $N$ candidates instead of a single answer~\cite{rupprecht2017learning}. Applied to human pose estimation it represents the conditional distribution $p(\mathbf{y}|\mathbf{x})$ over poses $\mathbf{y} \in \mathbb{R}^{K \times 3}$ with $K$ keypoints. It has been used mostly for monocular RGB pose estimation, where lifting ordered 2D keypoints to 3D leaves an irreducible depth ambiguity~\cite{zheng2023deep} that a deterministic model cannot resolve and that motivated multi-hypothesis RGB methods~\cite{wehrbein2021probabilistic, holmquist2023diffpose, gong2023diffpose, feng2023diffpose}. Early methods picked among hypotheses with best-of-$N$ losses and the minimum Mean Per Joint Position Error~\cite{wehrbein2021probabilistic, rupprecht2017learning}, which needs an oracle at inference and is impractical~\cite{pierzchlewicz2022multi, shan2023diffusion}, so recent methods model the posterior directly and aggregate its samples into a final estimate~\cite{11017419, jinwei2023diffpose, holmquist2023diffpose, feng2023diffpose, gong2023diffpose, li2021human}. These RGB methods fall into three classes, namely Gaussian models~\cite{li2019generating}, diffusion models~\cite{holmquist2023diffpose}, and normalizing flows~\cite{wehrbein2021probabilistic}.

Radar point clouds carry ambiguity that arises from sparsity, dropped detections, and multipath rather than depth, yet MHMs for radar remain largely under-explored. The few diffusion-based attempts, by Li~\etal~\cite{11017419, li2024rpdiff} and Fan~\etal~\cite{fan2024diffusion}, can sample multiple pose hypotheses but report only point predictions and quantify no uncertainty.

\subsection{Uncertainty for Radar Pose Estimation}
Calibration has received almost no attention in radar pose estimation. Chiang~\etal~\cite{chiang2024enhancing} reduce uncertainty during training but do not quantify uncertainty, and only Mueller~\etal~\cite{11729073} report calibrated uncertainty, fitting a heteroscedastic multivariate Gaussian on raw complex radar tensors and recalibrating it post-hoc on a held-out set. We compare against the radar Gaussian of Mueller~\etal, the radar diffusion model mmDiff of Fan~\etal~\cite{fan2024diffusion} evaluated in its stochastic multi-hypothesis form, and the canonical RGB diffusion model DiffPose~\cite{holmquist2023diffpose} adapted to radar.

\subsection{Normalizing Flows}
NFs learn invertible maps between a simple base distribution and a complex data distribution~\cite{papamakarios2021normalizing, pmlr-v267-zhai25d}, which yields exact likelihoods and efficient sampling. Conditional NFs have been used for probabilistic regression~\cite{winkler2019learning, pierzchlewicz2022multi} and for pose estimation, including multi-hypothesis 2D-to-3D lifting~\cite{wehrbein2021probabilistic, han2025propose}, weakly supervised reconstruction~\cite{10.1007/978-3-030-58539-6_28}, and anomaly detection~\cite{hirschorn2023normalizing}. Unlike continuous NFs that integrate an ODE over many steps~\cite{10.1007/978-3-031-95911-0_20}, discrete NFs use a fixed set of layers~\cite{dinh2017density}. To our knowledge, we are the first to apply conditional NFs to calibrated radar point-cloud pose estimation.

\section{Methods}
\label{sec:methods}
\begin{figure*}[t]
    \centering
    \includegraphics[width=\textwidth]{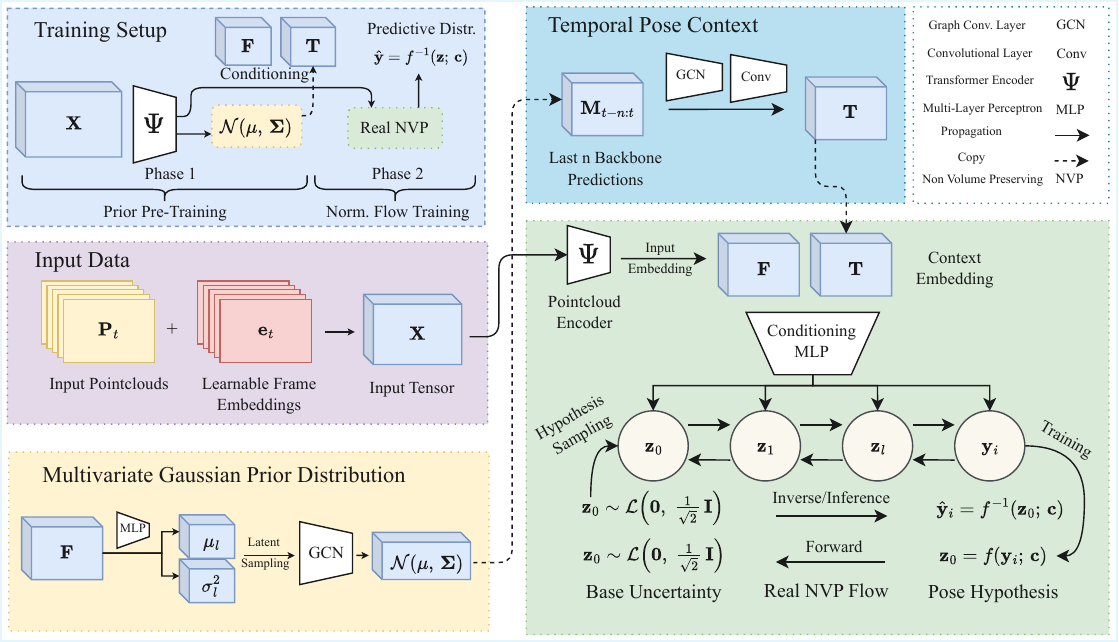}
    \caption{Overview of MH-NFPG. Phase~1 trains a transformer encoder on radar point cloud sequences to parameterize a heteroscedastic Gaussian prior over 3D poses. Phase~2 freezes the backbone and prior and trains a conditional Real NVP NF that transforms a Laplace base distribution into expressive pose hypotheses, conditioned on backbone features and temporal predictions from Phase~1. At inference, hypotheses are generated by sampling from the base distribution and applying the inverse NF.}
    \label{fig:architecture}
\end{figure*}
We propose a two-phase training pipeline (Figure \ref{fig:architecture}) that combines a
multivariate Gaussian prior with a conditional NF
to produce calibrated posteriors over 3D poses. In the
first phase, we train the prior distribution, a spatiotemporal transformer that encodes an input sequence of
radar point clouds into a latent distribution, from which samples are
decoded into 3D pose hypotheses by a graph convolutional network (GCN)
reflecting the kinematic structure of the human body.

We then use a conditional bijective NF in the second phase to predict the final output distribution. The NF starts from a standard Laplace distribution and is conditioned on two complementary
signals, the prior latent features providing per-frame spatial
context, and a temporal context vector from a second GCN encoding the
past $n_t$ prior predictions to enforce temporal consistency.

\subsection{Multi-Hypothesis Normalizing Flow Pose Generator}
\noindent\textbf{Spatiotemporal Transformer Backbone}
A radar sequence consists of $T$ frames $\mathcal{P} = \{\mathbf{P}_1, \ldots, \mathbf{P}_T\}$, where each frame $\mathbf{P}_t = \{\mathbf{x}_{t,1}, \ldots, \mathbf{x}_{t,N_t}\}$ is an unordered point cloud whose size $N_t$ varies across frames as CFAR retains a different number of detections. Each point is first embedded into a $d$-dimensional feature space via an MLP
$f_{\mathrm{emb}}: \mathbb{R}^{d_{\text{in}}} \rightarrow \mathbb{R}^d$.
A learnable temporal embedding $\mathbf{e}_t$ is then added to the embedded
points of each frame $t$, yielding $\mathbf{P}_t + \mathbf{e}_t$.
Finally, all $T$ frames are concatenated into a single point set
$\mathbf{X} \in \mathbb{R}^{(T \cdot N_t) \times d}$, which serves as input
to the transformer. Our backbone is a transformer encoder~\cite{vaswani2017attention} $\Psi(\cdot)$ that processes $\mathbf{X}$ using multi-head self-attention (MHSA). Since we do not use positional encodings for individual points, the spatial ordering within each frame is irrelevant, exploiting the permutation-equivariance of MHSA. The temporal embeddings, however, enable the model to distinguish points from different timesteps, preserving temporal structure. A learnable class token aggregates the sequence into a fixed-dimensional representation $\mathbf{F} \in \mathbb{R}^{d_F}$, achieving permutation invariance over the input point cloud.

\noindent\textbf{Prior Distribution}
Radar signals are inherently ambiguous due to sensor noise, multipath propagation and scattering~\cite{richards2005fundamentals}. We model this aleatoric uncertainty with a heteroscedastic multivariate Gaussian (MG) prior distribution~\cite{gundavarapu2019structured, 11729073}, which the NF uses as a conditioning signal to predict a more expressive posterior (Figure~\ref{fig:architecture}). The MG prior captures inter-keypoint correlations through the multivariate Gaussian likelihood, which optimizes the full covariance structure of the decoded pose hypotheses.

To achieve this, the backbone representation $\mathbf{F}$ is fed into two branches predicting mean $\boldsymbol{\mu}_l$ and log-variance $\log \boldsymbol{\sigma}_l^2$ for a $d_l$-dimensional latent Gaussian, regularized with a KL divergence term
\begin{equation}
    \mathcal{L}_{\text{KL}} = -\frac{1}{2d_l} \mathbb{E}\left[\sum_{j=1}^{d_l} \left(1 + \log \boldsymbol{\sigma}_{l,j}^2 - \boldsymbol{\mu}_{l,j}^2 - \boldsymbol{\sigma}_{l,j}^2\right)\right].
\end{equation}
We sample $N$ hypotheses and decode each with a spectral GCN~\cite{fan2024diffusion, gong2023diffpose} to predict the pose hypothesis distribution $\mathbf{E} \in \mathbb{R}^{3K \times N}$. The empirical mean $\boldsymbol{\mu}$ and covariance $\boldsymbol{\Sigma}$ of the decoded hypotheses are used in the multivariate negative log likelihood (NLL) loss. We use a variance regularizer $\mathcal{R} = \frac{\operatorname{tr}(\boldsymbol{\Sigma})}{d_{\text{pose}}}$ scaled by a hyperparameter $\gamma$ penalizing the mean diagonal variance of the covariance matrix to stabilize training.
\begin{equation}
\mathcal{L}_{\text{NLL}}^{\text{prior}} = \mathbb{E}\left[\frac{1}{2}\left(\log|\boldsymbol{\Sigma}| + (\mathbf{y} - \boldsymbol{\mu})^\top \boldsymbol{\Sigma}^{-1} (\mathbf{y} - \boldsymbol{\mu})\right) + \gamma \ \mathcal{R}\right].
\end{equation}
The final training loss for the prior is
\begin{equation}
     \mathcal{L}_{\text{backbone}} = \mathcal{L}_{\text{NLL}}^{\text{prior}} + \lambda_{\text{KL}}\mathcal{L}_{\text{KL}}.
\end{equation}
We additionally test Laplace prior distribution formulations (see the supplementary material) but achieve superior performance with the MG formulation. We tune hyperparameters $\gamma$ and $\lambda_{KL}$ with targeted grid search on a validation set (see the supplementary material).

\subsubsection{Conditional Normalizing Flow}
While Gaussians provide tractable uncertainty, their parametric assumptions may not capture the true uncertainty structure~\cite{wehrbein2021probabilistic, han2025propose}. We therefore use a conditional NF to transform a Laplace base distribution~\cite{li2021human} into a more expressive posterior conditioned on our prior. Formally, the NF models pose $\mathbf{y} \in \mathbb{R}^{d_{\text{pose}}}$, where $d_{\text{pose}} = 3K$ represents the flattened 3D coordinates of $K$ keypoints. The NF transforms samples from a base distribution of independent $\text{Laplace}(0,\; 1/\sqrt{2})$ components, chosen such that each marginal has unit variance, into an expressive pose distribution through a series of invertible transformations. We ablate further base distribution assumptions to validate this choice in Section \ref{ablation_study}.

\noindent\textbf{Affine Coupling Layers}
Following prior multi-hypothesis RGB pose estimation work~\cite{wehrbein2021probabilistic, li2021human}, we use Real NVP affine coupling layers~\cite{dinh2017density}, which allow efficient parallel sampling. The NF consists of $L$ such layers. Each layer $\ell$ applies a bijective transformation $f_\ell: \mathbb{R}^{d_{\text{pose}}} \rightarrow \mathbb{R}^{d_{\text{pose}}}$ using a binary mask $\mathbf{m}_\ell \in \{0,1\}^{d_{\text{pose}}}$ that alternates between even and odd dimensions across layers. The forward transformation is defined as
\begin{equation}
    \mathbf{h}_\ell = \mathbf{m}_\ell \odot \mathbf{h}_{\ell-1} + (1 - \mathbf{m}_\ell) \odot \left(\mathbf{h}_{\ell-1} \odot \exp(\mathbf{s}_\ell) + \mathbf{t}_\ell\right),
\end{equation}
where $\mathbf{h}_0 = \mathbf{y}$ is the input pose and $\odot$ denotes element-wise multiplication. The scale $\mathbf{s}_\ell \in \mathbb{R}^{d_{\text{pose}}}$ and translation $\mathbf{t}_\ell \in \mathbb{R}^{d_{\text{pose}}}$ parameters are computed by a neural network $g_\ell$ that takes both the masked input and the conditioning vector:
\begin{equation}
    [\tilde{\mathbf{s}}_\ell, \mathbf{t}_\ell] = g_\ell\!\left(\left[\mathbf{m}_\ell \odot \mathbf{h}_{\ell-1};\; \mathbf{c}\right]\right), \quad \mathbf{s}_\ell = \tanh(\tilde{\mathbf{s}}_\ell).
\end{equation}
Each $g_\ell$ is a three-layer MLP with dropout after each layer.

\noindent\textbf{Conditioning Mechanism}
We condition the model on two components, namely the backbone features $\mathbf{F} \in \mathbb{R}^{d_F}$ and the past temporal pose predictions $\mathbf{M} \in \mathbb{R}^{n_t \times K \times 3}$ of the prior distribution, where $d_F$ is the backbone feature dimension and $n_t$ the number of past frames. These temporal predictions supply motion context that compensates for frames where low-cross-section joints such as hands and feet drop out of the radar point cloud.
To capture biomechanical dependencies between skeletal keypoints, we employ a GCN for conditioning. Backbone features $\mathbf{F}$ are first projected to per-keypoint embeddings $\mathbf{P} \in \mathbb{R}^{K \times d_g}$ via a two-layer MLP, where $d_g$ is the per-keypoint embedding dimension. These are then processed through a Chebyshev graph convolution on the skeleton adjacency matrix to obtain a global embedding $\mathbf{G} \in \mathbb{R}^{K \times d_h}$, where $d_h$ is the GCN hidden dimension~\cite{fan2024diffusion, gong2023diffpose}. Similarly, each of the $n_t$ past pose frames in $\mathbf{M}$ is independently processed through a graph convolution with shared weights, producing per-frame embeddings in $\mathbb{R}^{K \times d_h}$. The resulting temporal sequence is aggregated via 1D convolution and max-pooling to form $\mathbf{T} \in \mathbb{R}^{K \times d_h}$. The global and temporal embeddings are summed element-wise, flattened to $\mathbb{R}^{K \cdot d_h}$, and projected through an MLP to produce the final conditioning vector $\mathbf{c} \in \mathbb{R}^{d_c}$, where $d_c$ is the conditioning dimension. In each affine coupling layer of the NF, $\mathbf{c}$ is concatenated with the masked input before passing through the scale-translation network.

\noindent\textbf{NF Composition}
The complete NF $\phi$ composes all $L$ affine coupling layers:
\begin{equation}
    \mathbf{z} = \phi(\mathbf{y}; \mathbf{c}) = f_L \circ f_{L-1} \circ \cdots \circ f_1(\mathbf{y}; \mathbf{c}).
\end{equation}
The inverse transformation $\mathbf{y} = \phi^{-1}(\mathbf{z}; \mathbf{c})$ maps from the base space $\mathbf{z} \in \mathbb{R}^{d_{\text{pose}}}$ back to pose space by applying the inverse of each coupling layer in reverse order. The log-likelihood of a pose $\mathbf{y}$ under the model is obtained via the change-of-variables formula~\cite{dinh2017density, ardizzone2018analyzing}
\begin{equation}
    \log p(\mathbf{y}|\mathbf{c}) = \log p_0(\mathbf{z}) + \log \left|\det \frac{\partial \mathbf{z}}{\partial \mathbf{y}}\right|,\; \text{where } \mathbf{z} = \phi(\mathbf{y}; \mathbf{c}).
\end{equation}

These components together form our \textbf{M}ulti-\textbf{H}ypothesis \textbf{N}ormalizing \textbf{F}low \textbf{P}ose \textbf{G}enerator model (MH-NFPG), which we report as MG-Prior + NFPG in our tables to make the conditioning prior explicit alongside the baselines.

\noindent\textbf{NF Training}
In the forward direction, ground-truth poses $\mathbf{y}^*$ are mapped to the base distribution via $\mathbf{z} = \phi(\mathbf{y}^*; \mathbf{c})$, and we minimize the negative log-likelihood
\begin{equation}
    \mathcal{L}_{\text{NLL}}^{\text{NF}} = -\mathbb{E}\left[\log p_0(\mathbf{z}) + \log \left|\det \frac{\partial \mathbf{z}}{\partial \mathbf{y}^*}\right|\right].
\end{equation}
The joint log-density of the base distribution $p_0$ with scale $b = 1/\sqrt{2}$ factorizes as
\begin{equation}
    \log p_0(\mathbf{z}) = d_{\text{pose}} \log \frac{1}{2b} - \frac{1}{b} \|\mathbf{z}\|_1.
\end{equation}
Since $\phi$ is a composition of $L$ affine coupling layers, each with a triangular Jacobian, the log-determinant decomposes as
\begin{equation}
    \log \left|\det \frac{\partial \mathbf{z}}{\partial \mathbf{y}^*}\right| = \sum_{\ell=1}^{L} \sum_{q \in \mathcal{U}_\ell} s_{\ell,q},
\end{equation}
where $s_{\ell,q}$ is the log-scale output of layer $\ell$ for dimension $q$ and $\mathcal{U}_\ell = \{q : m_{\ell,q} = 0\}$ denotes the set of unmasked dimensions in layer $\ell$. For inference, we sample $\mathbf{z}^{(s)} \sim p_0(\mathbf{z})$ and generate pose predictions via the inverse NF $\hat{\mathbf{y}}^{(s)} = \phi^{-1}(\mathbf{z}^{(s)}; \mathbf{c})$.

\subsection{Datasets}

We focus on compact frequency-modulated continuous-wave (FMCW) radars with 3Tx/4Rx antenna arrays, which enable fast on-device CFAR-based point cloud extraction~\cite{10904474, NEURIPS2022_af9c9c6d, krauss2024review}. While large-aperture arrays~\cite{chen2022mmbody, fan2024diffusion} achieve higher angular resolution, their substantially larger data volume~\cite{krauss2024review} prohibits real-time on-device processing, practical data transfer and hardware costs for applications. In line with prior radar pose estimation work~\cite{fan2024diffusion, NEURIPS2022_af9c9c6d, 10904474, 9083948, lee2023hupr, ho2024rt}, we evaluate on three publicly available single-person datasets. To ensure point cloud density, each input frame merges the points of the current and the four preceding radar frames~\cite{fan2024diffusion}, and the model takes a sequence of five consecutive such frames as input.

\noindent\textbf{MM-Fi Dataset}~\cite{yang2024mmfi}: This dataset comprises over 320k frames from 40 subjects (11 female, 29 male) performing 27 daily and rehabilitation activities across four environments. Ground-truth 17-keypoint poses are obtained via HRNet-w48 2D detection from two infrared cameras, triangulation, and optimization-based refinement. We follow the splits of Fan~\etal~\cite{fan2024diffusion} but use all available activities to increase dataset size and generalization.

\noindent\textbf{mmRadPose Dataset}~\cite{10904474}: This dataset comprises recordings of 12 participants performing 11 rehabilitation exercises with 26-keypoint annotations. Data were captured from three distinct radar aspect angles ($0^\circ$, $45^\circ$, and $90^\circ$ relative to the subject's frontal plane), providing viewpoint diversity for evaluating pose estimation robustness. The held-out test set contains one male and one female participant, evaluated across all exercises. Ground-truth poses are recorded with an optical motion capture system, yielding the highest-fidelity annotations among the three datasets.

\noindent\textbf{mRI Dataset}~\cite{NEURIPS2022_af9c9c6d}: This dataset includes 20 participants performing 12 distinct activities with 17-keypoint annotations. Ten activities consist of structured rehabilitation exercises, while the remaining two capture free-form stretching and relaxation movements, as well as straight-line walking. This broader activity set enables assessment of generalization across varying motion patterns. We use all exercises and 16 participants for training and 4 for testing. Ground-truth poses are obtained using 2D keypoint detection with HRNet from two RGB cameras, followed by triangulation and optimization-based refinement. The ground truth annotations contain occasional glitches, where poses jump around 2 meters between frames. We filter out those confounding frames for training (1\%) and testing (0.9\%).

\begin{table*}[t]
\centering
\begin{minipage}[t]{0.49\textwidth}
\centering
\caption{Pose estimation accuracy and uncertainty calibration on MM-Fi with standard testing protocols. Position errors in cm. @$X$\% denotes the proportion of ground-truth joints within the $X$\% credible interval.}
\label{tab:mmfi_results}
\resizebox{\linewidth}{!}{%
\begin{tabular}{l | cc | ccccc}
\toprule
\textbf{Model} & \textbf{MPJPE}$\downarrow$ & \textbf{PA-MPJPE}$\downarrow$ & \textbf{@50\%} & \textbf{@90\%} & \textbf{@95\%} & \textbf{ECE}$\downarrow$ & \textbf{Sharp.} \\
\midrule
\multicolumn{8}{c}{\textit{Random}} \\
\midrule
PointTransformer~\cite{zhao2021point} & 6.697 & 5.204 & -- & -- & -- & -- & -- \\
mmDiff ($\eta{=}0$)~\cite{fan2024diffusion} & 6.504 & 5.005 & -- & -- & -- & -- & -- \\
mmDiff ($\eta{=}1$)~\cite{fan2024diffusion} & 6.508 & 5.010 & 0.257 & 0.582 & 0.651 & 0.119 & 1.780 \\
\midrule
MG-Prior (Ours) & 6.533 & 4.937 & \textbf{0.447} & 0.763 & 0.823 & 0.040 & 2.901 \\
MG-Prior + DiffPose~\cite{holmquist2023diffpose} & 6.506 & 4.872 & 0.354 & 0.698 & 0.760 & 0.071 & 3.014 \\
MG-Prior + NFPG (Ours) & \textbf{6.206} & \textbf{4.687} & 0.583 & \textbf{0.902} & \textbf{0.943} & \textbf{0.038} & 5.457 \\
\midrule
\multicolumn{8}{c}{\textit{Cross-Subject}} \\
\midrule
PointTransformer~\cite{zhao2021point} & 6.665 & 5.092 & -- & -- & -- & -- & -- \\
mmDiff ($\eta{=}0$)~\cite{fan2024diffusion} & 6.363 & 4.901 & -- & -- & -- & -- & -- \\
mmDiff ($\eta{=}1$)~\cite{fan2024diffusion} & 6.360 & 4.904 & 0.245 & 0.565 & 0.637 & 0.122 & 1.719 \\
\midrule
MG-Prior (Ours) & 6.313 & 4.841 & \textbf{0.428} & 0.762 & 0.824 & 0.050 & 2.927 \\
MG-Prior + DiffPose~\cite{holmquist2023diffpose} & 6.359 & 4.787 & 0.342 & 0.684 & 0.746 & 0.078 & 2.932 \\
MG-Prior + NFPG (Ours) & \textbf{6.090} & \textbf{4.660} & 0.604 & \textbf{0.909} & \textbf{0.946} & \textbf{0.046} & 5.738 \\
\midrule
\multicolumn{8}{c}{\textit{Cross-Environment}} \\
\midrule
PointTransformer~\cite{zhao2021point} & 8.695 & 6.777 & -- & -- & -- & -- & -- \\
mmDiff ($\eta{=}0$)~\cite{fan2024diffusion} & 8.611 & 6.643 & -- & -- & -- & -- & -- \\
mmDiff ($\eta{=}1$)~\cite{fan2024diffusion} & 8.609 & 6.646 & 0.184 & 0.446 & 0.511 & 0.160 & 1.619 \\
\midrule
MG-Prior (Ours) & 8.780 & 6.770 & 0.392 & 0.687 & 0.747 & 0.065 & 3.028 \\
MG-Prior + DiffPose~\cite{holmquist2023diffpose} & 8.693 & 6.682 & 0.281 & 0.588 & 0.650 & 0.109 & 2.971 \\
MG-Prior + NFPG (Ours) & \textbf{8.187} & \textbf{6.312} & \textbf{0.533} & \textbf{0.852} & \textbf{0.901} & \textbf{0.025} & 5.841 \\
\bottomrule
\end{tabular}%
}
\end{minipage}
\hfill
\begin{minipage}[t]{0.49\textwidth}
\centering
\footnotesize
\caption{Leave-one-environment-out cross-validation on MM-Fi. Position errors in cm. Each block holds out one environment for testing and trains on the rest. Holding out Env~4 corresponds to the Cross-Environment split of Tab.~\ref{tab:mmfi_results}.}
\label{tab:loeo}
\resizebox{\linewidth}{!}{%
\begin{tabular}{ll|cc|ccccc}
\toprule
\textbf{Left-out} & \textbf{Model} & \textbf{MPJPE}$\downarrow$ & \textbf{PA-MPJPE}$\downarrow$ & \textbf{@50\%} & \textbf{@90\%} & \textbf{@95\%} & \textbf{ECE}$\downarrow$ & \textbf{Sharp.} \\
\midrule
\multirow{6}{*}{Env 1} & PointTransformer~\cite{zhao2021point} & 7.124 & 5.676 & -- & -- & -- & -- & -- \\
 & mmDiff ($\eta{=}0$)~\cite{fan2024diffusion} & 6.941 & 5.530 & -- & -- & -- & -- & -- \\
 & mmDiff ($\eta{=}1$)~\cite{fan2024diffusion} & 6.936 & 5.532 & 0.232 & 0.535 & 0.605 & 0.131 & 1.668 \\
 & MG-Prior (Ours) & \textbf{6.699} & \textbf{5.238} & \textbf{0.437} & 0.747 & 0.807 & 0.044 & 2.818 \\
 & MG-Prior + DiffPose~\cite{holmquist2023diffpose} & 6.894 & 5.339 & 0.366 & 0.718 & 0.782 & 0.065 & 3.367 \\
 & MG-Prior + NFPG (Ours) & 6.744 & 5.329 & 0.601 & \textbf{0.911} & \textbf{0.947} & \textbf{0.043} & 6.294 \\
\midrule
\multirow{6}{*}{Env 2} & PointTransformer~\cite{zhao2021point} & 7.169 & 5.675 & -- & -- & -- & -- & -- \\
 & mmDiff ($\eta{=}0$)~\cite{fan2024diffusion} & 7.033 & 5.505 & -- & -- & -- & -- & -- \\
 & mmDiff ($\eta{=}1$)~\cite{fan2024diffusion} & 7.044 & 5.515 & 0.228 & 0.533 & 0.607 & 0.132 & 1.721 \\
 & MG-Prior (Ours) & 7.021 & 5.440 & 0.390 & 0.706 & 0.771 & 0.063 & 2.755 \\
 & MG-Prior + DiffPose~\cite{holmquist2023diffpose} & 7.276 & 5.643 & 0.359 & 0.698 & 0.761 & 0.071 & 3.364 \\
 & MG-Prior + NFPG (Ours) & \textbf{6.813} & \textbf{5.326} & \textbf{0.599} & \textbf{0.908} & \textbf{0.943} & \textbf{0.042} & 6.091 \\
\midrule
\multirow{6}{*}{Env 3} & PointTransformer~\cite{zhao2021point} & 7.012 & 5.506 & -- & -- & -- & -- & -- \\
 & mmDiff ($\eta{=}0$)~\cite{fan2024diffusion} & 6.776 & 5.298 & -- & -- & -- & -- & -- \\
 & mmDiff ($\eta{=}1$)~\cite{fan2024diffusion} & 6.778 & 5.301 & 0.254 & 0.566 & 0.631 & 0.119 & 1.713 \\
 & MG-Prior (Ours) & 6.716 & 5.220 & 0.420 & 0.736 & 0.797 & 0.050 & 2.744 \\
 & MG-Prior + DiffPose~\cite{holmquist2023diffpose} & 6.914 & 5.245 & 0.365 & 0.712 & 0.776 & 0.068 & 3.300 \\
 & MG-Prior + NFPG (Ours) & \textbf{6.382} & \textbf{4.899} & \textbf{0.576} & \textbf{0.890} & \textbf{0.931} & \textbf{0.037} & 5.382 \\
\midrule
\multirow{6}{*}{Env 4} & PointTransformer~\cite{zhao2021point} & 8.695 & 6.777 & -- & -- & -- & -- & -- \\
 & mmDiff ($\eta{=}0$)~\cite{fan2024diffusion} & 8.611 & 6.643 & -- & -- & -- & -- & -- \\
 & mmDiff ($\eta{=}1$)~\cite{fan2024diffusion} & 8.609 & 6.646 & 0.184 & 0.446 & 0.511 & 0.160 & 1.619 \\
 & MG-Prior (Ours) & 8.780 & 6.770 & 0.392 & 0.687 & 0.747 & 0.065 & 3.028 \\
 & MG-Prior + DiffPose~\cite{holmquist2023diffpose} & 8.693 & 6.682 & 0.281 & 0.588 & 0.650 & 0.109 & 2.971 \\
 & MG-Prior + NFPG (Ours) & \textbf{8.187} & \textbf{6.312} & \textbf{0.533} & \textbf{0.852} & \textbf{0.901} & \textbf{0.025} & 5.841 \\
\bottomrule
\end{tabular}%
}
\end{minipage}
\end{table*}

\subsection{Evaluation Metrics}
We evaluate pose accuracy with Mean Per Joint Position Error (MPJPE), the average Euclidean distance between predicted and ground-truth keypoints, $\text{MPJPE} = \frac{1}{K}\sum_{k=1}^{K}\|\hat{\mathbf{y}}_k - \mathbf{y}^*_k\|_2$. We also report PA-MPJPE, which adds Procrustes alignment.

We assess uncertainty directly from the pose hypotheses. For each pose coordinate they form an empirical CDF $\hat{F}$, and evaluating it at the ground truth yields the probability integral transform $u_i = \hat{F}(y^*_i)$, the fraction of hypotheses below the true value, which is uniform on $[0,1]$ under perfect calibration. The Expected Calibration Error (ECE)~\cite{naeini2015obtaining, pierzchlewicz2022multi} measures its deviation from uniformity across $C{=}100$ levels $p_j \in [0.01, 0.99]$,
\begin{equation}
\text{ECE} = \sqrt{\frac{1}{C}\sum_{j=1}^{C} \left(\hat{P}(p_j) - p_j\right)^2},
\end{equation}
where $\hat{P}(p_j) = \frac{1}{N}\sum_{i=1}^{N}\mathbf{1}[u_i \leq p_j]$ and $N$ is the number of test samples. We use the L2 variant, which penalizes large deviations more heavily than the original L1 formulation~\cite{naeini2015obtaining}. We also report interval coverage at the 50\%, 90\%, and 95\% levels, the fraction of ground truths inside the central quantile interval $[\hat{F}^{-1}(\tfrac{1-c}{2}),\, \hat{F}^{-1}(\tfrac{1+c}{2})]$ for level $c$, and sharpness, the average predictive standard deviation,
\begin{equation}
\text{Sharpness} = \frac{1}{NK}\sum_{i=1}^{N}\sum_{k=1}^{K}\sigma_{i,k}.
\end{equation}

\begin{table*}[b]
\centering
\begin{minipage}[t]{0.49\textwidth}
\centering
\caption{Pose estimation accuracy and uncertainty calibration on mmRadPose. Position errors in cm. Per-angle columns report MPJPE only.}
\label{tab:mmradpose_results}
\resizebox{\linewidth}{!}{%
\begin{tabular}{l | cc | ccc | ccccc}
\toprule
\textbf{Model}
& \multicolumn{2}{c|}{\textbf{Overall}}
& \multicolumn{3}{c|}{\textbf{MPJPE by Angle}}
& \multicolumn{5}{c}{\textbf{Calibration (Overall)}} \\
& \textbf{MPJPE}$\downarrow$ & \textbf{PA-MPJPE}$\downarrow$
& \textbf{0\degree} & \textbf{45\degree} & \textbf{90\degree}
& \textbf{@50\%} & \textbf{@90\%} & \textbf{@95\%} & \textbf{ECE}$\downarrow$ & \textbf{Sharp.} \\
\midrule
PointTransformer~\cite{zhao2021point}
& 6.413 & 4.920
& 5.958 & 6.232 & 7.072
& -- & -- & -- & -- & -- \\
mmDiff ($\eta{=}0$)~\cite{fan2024diffusion}
& 6.004 & 4.634
& 5.566 & 5.737 & 6.731
& -- & -- & -- & -- & -- \\
mmDiff ($\eta{=}1$)~\cite{fan2024diffusion}
& \textbf{5.992} & 4.634
& \textbf{5.555} & 5.720 & 6.723
& 0.203 & 0.488 & 0.560 & 0.166 & 1.546 \\
\midrule
MG-Prior (Ours)
& 6.645 & 4.933
& 6.565 & 6.252 & 7.121
& 0.346 & 0.698 & 0.773 & 0.109 & 2.755 \\
MG-Prior + DiffPose~\cite{holmquist2023diffpose}
& 6.539 & 4.812
& 6.331 & 6.176 & 7.120
& 0.254 & 0.545 & 0.613 & 0.145 & 2.497 \\
MG-Prior + NFPG (Ours)
& 6.030 & \textbf{4.597}
& 5.895 & \textbf{5.587} & \textbf{6.613}
& \textbf{0.480} & \textbf{0.839} & \textbf{0.891} & \textbf{0.056} & 4.739 \\
\bottomrule
\end{tabular}%
}
\end{minipage}
\hfill
\begin{minipage}[t]{0.49\textwidth}
\centering
\caption{Pose estimation accuracy and uncertainty calibration on mRI. Position errors in cm.}
\label{tab:mri_metrics}
\resizebox{\linewidth}{!}{%
\begin{tabular}{l | cc | ccccc}
\toprule
\textbf{Model} & \textbf{MPJPE}$\downarrow$ & \textbf{PA-MPJPE}$\downarrow$ & \textbf{@50\%} & \textbf{@90\%} & \textbf{@95\%} & \textbf{ECE}$\downarrow$ & \textbf{Sharp.} \\
\midrule
PointTransformer~\cite{zhao2021point} & 8.547 & 6.196 & -- & -- & -- & -- & -- \\
mmDiff ($\eta{=}0$)~\cite{fan2024diffusion} & 8.159 & 5.757 & -- & -- & -- & -- & -- \\
mmDiff ($\eta{=}1$)~\cite{fan2024diffusion} & \textbf{8.149} & 5.744 & 0.200 & 0.484 & 0.554 & 0.150 & 1.890 \\
\midrule
MG-Prior (Ours) & 8.371 & 5.808 & 0.303 & 0.635 & 0.709 & 0.099 & 2.796 \\
MG-Prior + DiffPose~\cite{holmquist2023diffpose} & 8.802 & 6.043 & 0.245 & 0.528 & 0.589 & 0.132 & 3.351 \\
MG-Prior + NFPG (Ours) & 8.269 & \textbf{5.616} & \textbf{0.477} & \textbf{0.829} & \textbf{0.888} & \textbf{0.038} & 6.175 \\
\bottomrule
\end{tabular}%
}
\end{minipage}
\end{table*}

\subsection{Baseline Models}
We evaluate our approach against baselines spanning deterministic and probabilistic paradigms using state-of-the-art models from the RGB and radar domain adapted to our use case.

\noindent\textbf{PointTransformer}~\cite{zhao2021point}\textbf{:} A deterministic point cloud processing baseline that directly regresses 3D poses without probabilistic modeling.

\noindent\textbf{mmDiff}~\cite{fan2024diffusion}\textbf{:} We implement the deterministic DDIM framework of Fan~\etal, which treats the backbone's coarse prediction as a noisy pose estimate and iteratively refines it through a deterministic diffusion process. While effective for accuracy, this formulation cannot quantify predictive ambiguity. Since a DDIM can equivalently be formulated as a DDPM by setting $\eta = 1$ in the denoising equation~\cite{song2021denoising}, we additionally evaluate this stochastic variant to enable multi-hypothesis generation and uncertainty quantification.

\noindent\textbf{MG-Prior:} We use our multi-hypothesis multivariate Gaussian model of Phase 1 as the prior of the downstream models and independently train it as a standalone model.

\noindent\textbf{MG-Prior + DiffPose}~\cite{holmquist2023diffpose}\textbf{:} We adapt the MHM DiffPose from the RGB domain to radar-based pose estimation by replacing its original heatmap-sample conditioning with samples from our MG prior distribution $\mathcal{N}(\boldsymbol{\mu}, \boldsymbol{\Sigma})$. In line with the original formulation, the $N$ sampled poses are embedded, weighted by their Gaussian likelihood, and aggregated via a joint-wise transformer. The output is concatenated with projected backbone features to form the conditioning vector. This baseline generates diverse pose hypotheses from standard Gaussian noise instead of a backbone prediction, enabling calibration and coverage analysis.

\subsection{Implementation Details}
\label{implementation}
Hyperparameters for the prior are tuned on the MM-Fi dataset using cross-subject evaluation with three held-out validation participants (see the supplementary material for full grid search results). We select hyperparameters that balance accuracy and calibration, yielding $\lambda_{\text{KL}}{=}15$ and $\gamma{=}1$ for the Gaussian prior distribution, which we use for all datasets without re-tuning. We use 5 consecutive radar frames as inputs to the model and 6 consecutive backbone predictions for the conditioning of the NF. The NF is implemented with 8 affine coupling layers~\cite{dinh2017density}. The prior and MH-NFPG are trained with the AdamW optimizer, a batch size of 32 and a learning rate of 0.0001 for both phases on NVIDIA A40 GPUs. We train the model using early stopping, resulting in 5 epochs of prior training and 10 epochs of NF training for the three MM-Fi conditions, 15 and 25 epochs on mmRadPose, and 19 and 6 epochs on the mRI dataset, respectively.

Baseline models are trained using the optimized parameters reported in their respective implementations~\cite{fan2024diffusion, holmquist2023diffpose}. For mmDiff, we retain the limb loss weight of 10 from the original code repository. All evaluations of probabilistic models are done with 200 hypotheses~\cite{holmquist2023diffpose, wehrbein2021probabilistic}. The final pose estimate is obtained by averaging across all hypotheses.

\section{Experiments}
\label{sec:results}
We evaluate MH-NFPG on three radar pose estimation benchmarks, MM-Fi~\cite{yang2024mmfi}, mmRadPose~\cite{10904474}, and mRI~\cite{NEURIPS2022_af9c9c6d}. Tables~\ref{tab:mmfi_results}--\ref{tab:mri_metrics} report pose estimation accuracy and uncertainty calibration jointly. We first analyze calibration, the primary contribution of this work, and then confirm that calibrated uncertainty does not sacrifice prediction accuracy. We subsequently evaluate inference efficiency and ablate key design choices.

\subsection{Uncertainty Calibration and Pose Estimation}

Table~\ref{tab:mmfi_results} reports MM-Fi under its official testing protocol, the Random, Cross-Subject, and Cross-Environment splits, and Table~\ref{tab:loeo} additionally evaluates a leave-one-environment-out cross-validation that holds out each environment in turn. MH-NFPG consistently produces well-calibrated posteriors across all three benchmarks. Across the two diffusion baselines, MH-NFPG reduces ECE by 34--84\%, with the largest reduction on the Cross-Environment split of MM-Fi (ECE 0.025 against 0.160 for mmDiff and 0.109 for MG-Prior + DiffPose), where distribution shift causes the diffusion baselines to degrade severely while our posterior stays calibrated. Coverage at 95\% closely tracks the nominal rate across all settings (89--95\%), whereas mmDiff ($\eta{=}1$) exhibits severe undercoverage (51--65\%). MG-Prior + DiffPose also falls short of nominal coverage, so neither diffusion baseline matches the calibration of our NF.

\begin{table}[b]
\centering
\caption{Computational comparison of multi-hypothesis pose estimation models on mmRadPose using 200 hypotheses for all probabilistic models. The bottom section shows the MH-NFPG component breakdown.}
\label{tab:computational_comparison}
\resizebox{\columnwidth}{!}{%
\begin{tabular}{lccc}
\toprule
\textbf{Model} & \textbf{Params (M)} & \textbf{GFLOPs} & \textbf{Time (ms)} \\
\midrule
MG-Prior & 10.609 & 27.889 & 6.235 $\pm$ 0.012 \\
MG-Prior + NFPG & 19.639 & 28.767 & 10.786 $\pm$ 0.054 \\
MG-Prior + DiffPose~\cite{holmquist2023diffpose} & 26.224 & 64.335 & 25.099 $\pm$ 0.123 \\
mmDiff~\cite{fan2024diffusion} $\eta{=}0$ (det., 1 hyp.) & 47.181 & 62.942 & 40.959 $\pm$ 0.235 \\
mmDiff~\cite{fan2024diffusion} $\eta{=}1$ & 47.181 & 281.814 & 236.602 $\pm$ 2.528 \\
\midrule
\midrule
\multicolumn{4}{l}{\textbf{MG-Prior + NFPG Component Breakdown}} \\
\midrule
Spatiotemporal Transformer Backbone & 8,945,152 & 27.720 & 4.752 $\pm$ 0.014 \\
Gaussian Head & 1,663,698 & 0.426 & 1.437 $\pm$ 0.011 \\
NF Conditioning Module & 7,496,800 & 0.010 & 2.367 $\pm$ 0.023 \\
Standard Laplace Sampling & --- & 0.000 & 0.139 $\pm$ 0.003 \\
Real NVP NF & 1,533,152 & 0.611 & 2.090 $\pm$ 0.021 \\
\midrule
\textbf{Total} & \textbf{19,638,802} & \textbf{28.767} & \textbf{10.786 $\pm$ 0.054} \\
\bottomrule
\end{tabular}%
}
\end{table}

\begin{figure}[t]
    \centering
    \includegraphics[width=\linewidth]{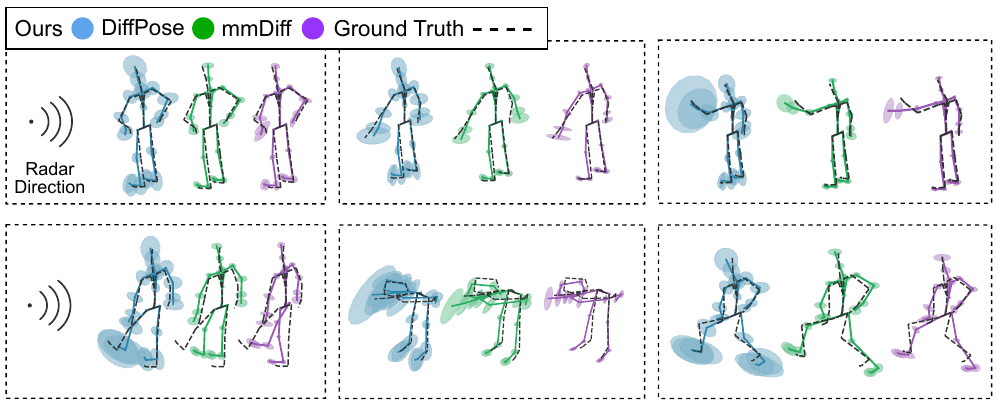}
    \caption{Qualitative comparison on mmRadPose. Predicted skeletons (solid, colored)
and ground truth (dotted) are shown with per-joint 90\% confidence ellipsoids. \textbf{Top}: Easy examples. Joints with smaller radar cross section show higher uncertainties. \textbf{Bottom}: Challenging examples
with occlusions and more complicated multipath effects. MH-NFPG achieves lower pose error with
well-calibrated uncertainty estimates, where ground-truth joints fall consistently
within the 90\% confidence regions. Baseline methods produce poorly localized ellipsoids
under ambiguous conditions.}
    \label{fig:qualitative_comparison}
\end{figure}

\begin{table*}[!b]
\centering
\caption{Ablation across all three datasets. Cells: MPJPE\,(cm)\,/\,ECE\,/\,Sharpness\,(cm). Base dist.: StdN=Std.\ Normal; ScG/ScL=Gaussian/Laplace scaled by the prior uncertainty. MM-Fi uses the cross-subject protocol. Default (bold): $L{=}8$, GCN, fixed-scale Laplace, affine, temporal.}
\label{tab:new_ablations}
\resizebox{\textwidth}{!}{%
\begin{tabular}{l|ccccc|c|ccc|c|c|c}
\toprule
 & \multicolumn{5}{c|}{Coupling layers} & Cond. & \multicolumn{3}{c|}{Base distribution} & NF & Temp. & \\
 & $L{=}2$ & $L{=}4$ & $L{=}6$ & $L{=}10$ & $L{=}12$ & MLP & StdN & ScG & ScL & Spline & None & \textbf{Ours} \\
\midrule
\textbf{MM-Fi} & 6.372/.056/5.582 & 6.260/.080/7.644 & 6.159/.060/6.341 & 6.253/.033/4.739 & 6.213/.026/4.400 & 6.186/.057/6.388 & 6.354/.032/4.849 & 6.369/.028/4.690 & 6.314/.036/5.349 & 6.361/.042/11.73 & 6.281/.049/5.829 & \textbf{6.090/.046/5.738} \\
\textbf{mmRadPose} & 6.150/.061/3.284 & 6.294/.061/5.794 & 6.282/.054/5.815 & 6.164/.083/3.913 & 6.215/.113/3.434 & 6.333/.062/4.941 & 6.109/.074/3.719 & 6.285/.072/4.171 & 6.248/.059/5.220 & 6.481/.128/10.52 & 6.408/.061/4.672 & \textbf{6.030/.056/4.739} \\
\textbf{mRI} & 8.979/.072/3.311 & 8.493/.024/7.038 & 8.442/.041/7.456 & 8.236/.060/4.205 & 8.246/.069/4.026 & 8.588/.033/7.028 & 8.288/.039/4.956 & 8.259/.074/4.421 & 8.275/.055/4.734 & 8.448/.088/11.70 & 8.397/.017/6.782 & \textbf{8.269/.038/6.175} \\
\bottomrule
\end{tabular}%
}
\end{table*}

Under this protocol, MH-NFPG stays well-calibrated across all four held-out environments (ECE 0.025--0.043) with the highest coverage at every credible level, whereas mmDiff ($\eta{=}1$) ranges from 0.119 to 0.160, a 3$\times$ to over 6$\times$ gap. MH-NFPG also attains the best position accuracy in three of the four splits, with the standalone MG prior marginally ahead on MPJPE and PA-MPJPE only when Env~1 is held out. This confirms that the learned posterior adapts to unseen environments rather than reflecting a single favorable split.

While MH-NFPG produces broader predictive intervals than baselines (reflected in higher sharpness values), the near-nominal coverage rates demonstrate that these intervals are appropriately sized rather than overconfident, which is critical for safety-relevant applications. In contrast, diffusion baselines exhibit low sharpness but severely undercover, indicating overconfident uncertainty estimates. Notably, MH-NFPG produces its tightest predictive intervals on mmRadPose, which uses optical motion capture ground truth, suggesting that the model's uncertainty appropriately reflects ground truth precision.

In pose accuracy MH-NFPG is on par with the strongest baseline on all three datasets. It attains the lowest MPJPE on every MM-Fi split and the lowest PA-MPJPE on mRI and mmRadPose (5.616 and 4.597~cm), where mmDiff is 0.12 and 0.04~cm ahead in MPJPE. Calibrated uncertainty therefore comes at no cost in point estimation accuracy.

\begin{figure}[t]
    \centering
    \includegraphics[width=\linewidth]{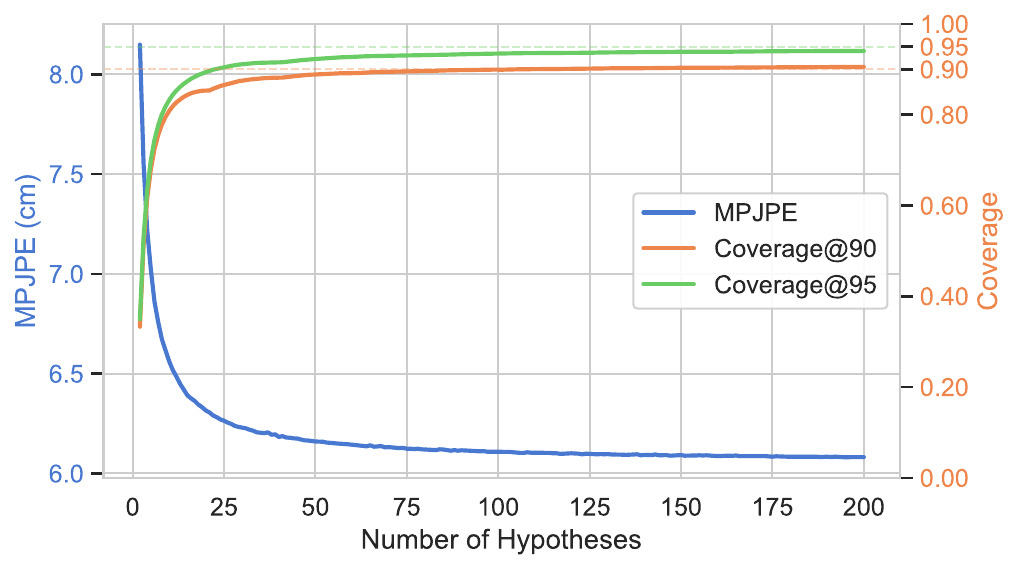}
    \caption{Effect of the number of hypotheses on MPJPE and calibration coverage on MM-Fi. Both improve with more hypotheses, with diminishing returns beyond approximately 100.}
    \label{fig:n_hypotheses}
\end{figure}

\subsection{Qualitative Analysis}

Figure~\ref{fig:qualitative_actions} shows MH-NFPG predictions across all three datasets, with ground truth consistently falling within the predicted confidence intervals, confirming that calibrated uncertainty generalizes across diverse actions, subjects, and radar configurations. Figure~\ref{fig:qualitative_comparison} compares MH-NFPG against DiffPose and mmDiff on mmRadPose. Our model produces well-localized estimates with confidence ellipsoids that reflect radar scattering properties~\cite{richards2005fundamentals}, especially under more complicated multipath effects and occlusions. In contrast, ground-truth joints frequently fall outside the predicted confidence ellipsoids for both MG-Prior + DiffPose and mmDiff. Further examples, including ambiguous cases with occlusions and multipath effects, are shown in the supplementary material.

\subsection{Inference Efficiency}

We evaluate all models in a causal, online setting with batch size 1 (see the supplementary material for details) and high-fidelity sampling from the posterior. For temporal models, we cache backbone predictions in a circular buffer, requiring only a single forward pass per frame. All 200 hypotheses \cite{holmquist2023diffpose} are drawn in parallel and stored in the batch dimension of the input tensor for all models, whereas diffusion baselines require additional sequential denoising steps (25 for DiffPose~\cite{holmquist2023diffpose} and mmDiff~\cite{fan2024diffusion} ($\eta{=}1$), 3 for mmDiff ($\eta{=}0$)).

Table~\ref{tab:computational_comparison} reports computational cost when generating high-fidelity multi-hypothesis distributions with 200 samples per frame on an NVIDIA RTX 3090 Ti GPU. MH-NFPG runs at 10.8~ms per frame (93~FPS), about 22$\times$ faster than mmDiff ($\eta{=}1$) and 2.3$\times$ faster than MG-Prior + DiffPose, with 10$\times$ fewer FLOPs than mmDiff ($\eta{=}1$). The efficiency gain stems from replacing sequential denoising with a single NF forward pass. The Real NVP NF requires only 2.1~ms.

\subsection{Ablation Study}
\label{ablation_study}

Table~\ref{tab:new_ablations} ablates the major design choices across all three datasets. The default configuration ($L{=}8$ coupling layers, GCN conditioning, Standard Laplace base, affine coupling, temporal context) gives the best overall accuracy/calibration trade-off on every dataset, attaining the lowest MPJPE on MM-Fi (6.090~cm) and mmRadPose (6.030~cm) and remaining on par with the best on mRI at a substantially lower ECE. Varying the coupling depth shows that deeper NFs ($L{\in}\{10,12\}$) increase MPJPE except on mRI, where they degrade calibration, whereas replacing the GCN conditioning with an MLP consistently worsens accuracy (e.g., 6.030 to 6.333~cm on mmRadPose), confirming the value of encoding skeletal structure. Among base distributions, the heavy-tailed Standard Laplace yields the best accuracy on MM-Fi and mmRadPose, supporting our choice. Substituting rational quadratic spline NFs~\cite{durkan2019neural} for affine coupling worsens accuracy and greatly inflates sharpness (up to 11.7~cm), and we omit autoregressive NFs due to their higher computational complexity~\cite{papamakarios2017masked}. Finally, removing the temporal conditioning increases MPJPE on all three datasets, confirming that temporal context disambiguates sparse radar observations. Figure~\ref{fig:n_hypotheses} further shows that both MPJPE and calibration coverage improve with increasing hypothesis count, with diminishing returns beyond approximately 100 hypotheses, validating our choice of 200.

\section{Conclusion}
\label{sec:conclusion}

We presented MH-NFPG, a normalizing flow pose generator for radar-based human pose estimation that combines probabilistic expressiveness with single forward-pass inference. A conditional NF with a heteroscedastic Gaussian prior and Laplace base distribution produces calibrated uncertainties while enabling parallel hypothesis sampling. On MM-Fi, mmRadPose, and mRI, MH-NFPG achieves the best uncertainty calibration at an accuracy on par with the strongest baselines, while enabling real-time deployment. Operating on radar point clouds at about 93 frames per second, it delivers the calibrated uncertainty that a safety-critical system needs to know when its prediction can be trusted.

\clearpage
{
    \small
    \bibliographystyle{ieeenat_fullname}
    \bibliography{references}
}

\appendix
\maketitlesupplementary

\section{Prior Loss Derivation}
\label{sec:prior_loss_derivation}
We test two formalizations of the prior loss.

\noindent\textbf{Gaussian Prior NLL.}
Starting from the multivariate Gaussian likelihood \cite{bramlage2023plausible}
\begin{equation}
p(\mathbf{y}) = \frac{1}{(2\pi)^{d/2}|\boldsymbol{\Sigma}|^{1/2}} \exp\!\left(-\frac{1}{2}(\mathbf{y} - \boldsymbol{\mu})^\top \boldsymbol{\Sigma}^{-1} (\mathbf{y} - \boldsymbol{\mu})\right),
\end{equation}
and dropping the constant normalization term, we obtain the negative log-likelihood. Adding a variance regularizer $\gamma\,\mathcal{R}$ with $\mathcal{R} = \frac{\operatorname{tr}(\boldsymbol{\Sigma})}{d_{\text{pose}}}$ to stabilize training yields
\begin{equation}
\mathcal{L}_{\text{NLL}}^{\text{prior}} = \mathbb{E}\left[\frac{1}{2}\left(\log|\boldsymbol{\Sigma}| + (\mathbf{y} - \boldsymbol{\mu})^\top \boldsymbol{\Sigma}^{-1} (\mathbf{y} - \boldsymbol{\mu})\right) + \gamma \ \mathcal{R}\right].
\end{equation}
In practice, the empirical mean $\boldsymbol{\mu}$ and covariance $\boldsymbol{\Sigma}$ are computed from the $N$ decoded pose hypotheses. To avoid explicit matrix inversion, we factorize $\boldsymbol{\Sigma} = \mathbf{L}\mathbf{L}^\top$ via Cholesky decomposition. The Mahalanobis distance is then obtained by solving the triangular system $\mathbf{L}\mathbf{v} = (\mathbf{y} - \boldsymbol{\mu})$ and computing $\|\mathbf{v}\|^2$, while the log-determinant reduces to $\log|\boldsymbol{\Sigma}| = 2\sum_i \log L_{ii}$. Adaptive diagonal jitter ensures numerical stability when $\boldsymbol{\Sigma}$ is near-singular.

\noindent\textbf{Laplace Prior NLL.}
Starting from the Laplace log-likelihood
\begin{equation}
\log p(\mathbf{y}) = \sum_{i=1}^{d_{\text{pose}}} \left(-\log(2b_i) - \frac{|y_i - \mu_i|}{b_i}\right),
\end{equation}
and dropping constants, we obtain the negative log-likelihood with scale regularizer $\mathcal{R} = \frac{\sum_{i=1}^{d_{\text{pose}}} b_i}{d_{\text{pose}}}$
\begin{equation}
\mathcal{L}_{\text{L1}} = \mathbb{E}\left[\sum_{i=1}^{d_{\text{pose}}} \left(\log b_i + \frac{|y_i - \mu_i|}{b_i}\right) + \gamma\, \mathcal{R}\right].
\end{equation}

\noindent\textbf{KL Divergence.}
The KL divergence between the learned latent distribution $q = \mathcal{N}(\boldsymbol{\mu}_l, \operatorname{diag}(\boldsymbol{\sigma}_l^2))$ and the standard normal prior $p = \mathcal{N}(\mathbf{0}, \mathbf{I})$ has a closed-form solution, which we average over the $d_l$ latent dimensions,
\begin{equation}
    \mathcal{L}_{\text{KL}} = -\frac{1}{2d_l} \mathbb{E}\left[\sum_{j=1}^{d_l} \left(1 + \log \boldsymbol{\sigma}_{l,j}^2 - \boldsymbol{\mu}_{l,j}^2 - \boldsymbol{\sigma}_{l,j}^2\right)\right].
\end{equation}

The final training loss for the prior combines the reconstruction and regularization terms
\begin{equation}
     \mathcal{L}_{\text{backbone}} = \mathcal{L}_{\text{NLL}}^{\text{prior}} + \lambda_{\text{KL}}\mathcal{L}_{\text{KL}}.
\end{equation}

\section{Hyperparameter Tuning}
\label{sec:hyperparameter_tuning}

We tune hyperparameters of the prior distribution on the MM-Fi dataset with cross-subject evaluation using three random held-out train-set participants as validation. Tables~\ref{tab:base_ablation} and~\ref{tab:kl_recon_ablation} present the full grid search results.

\begin{table*}
\centering
\caption{Tuning of base distribution hyperparameters $\lambda_{\text{KL}}$ (KL weight) and $\gamma$ (variance regularization) with Gaussian likelihood. MPJPE in cm, Sharpness in cm.}
\label{tab:base_ablation}
\resizebox{\textwidth}{!}{%
\begin{tabular}{l|ccccc|ccccc|ccccc|ccccc|ccccc}
\toprule
& \multicolumn{5}{c|}{$\lambda_{\text{KL}}{=}1$} & \multicolumn{5}{c|}{$\lambda_{\text{KL}}{=}5$} & \multicolumn{5}{c|}{$\lambda_{\text{KL}}{=}10$} & \multicolumn{5}{c|}{$\lambda_{\text{KL}}{=}15$} & \multicolumn{5}{c}{$\lambda_{\text{KL}}{=}20$} \\
$\gamma$ & 1 & 3 & 5 & 7 & 9 & 1 & 3 & 5 & 7 & 9 & 1 & 3 & 5 & 7 & 9 & 1 & 3 & 5 & 7 & 9 & 1 & 3 & 5 & 7 & 9 \\
\midrule
MPJPE$\downarrow$ & 6.71 & 6.86 & \textbf{6.60} & 6.78 & 6.68 & 6.73 & 6.67 & 6.77 & 6.68 & 6.64 & 6.74 & 6.72 & 6.78 & 6.73 & 6.73 & \underline{6.67} & 6.74 & 6.70 & 6.73 & 6.84 & 6.87 & 6.76 & 7.01 & 6.73 & 6.66 \\
ECE$\downarrow$ & .044 & .069 & .079 & .094 & .095 & .052 & .072 & .085 & .090 & .092 & .051 & .067 & .085 & .090 & .094 & \underline{.042} & .073 & .083 & .100 & .093 & .047 & .070 & \textbf{.040} & .085 & .098 \\
Sharpness & 8.26 & 5.26 & 4.06 & 3.59 & 3.24 & 7.77 & 5.03 & 3.81 & 3.46 & 3.21 & 7.74 & 4.95 & 4.05 & 3.40 & 3.12 & 8.41 & 4.90 & 4.06 & 3.38 & 3.10 & 8.11 & 4.93 & 7.93 & 3.44 & 3.25 \\
\bottomrule
\end{tabular}%
}
\\[2pt]
{\footnotesize Best values in \textbf{bold}, \underline{underlined} denotes best accuracy-calibration trade-off ($\lambda_{\text{KL}}{=}15$, $\gamma{=}1$). Evaluated on MM-Fi cross-subject split with three validation participants.}
\end{table*}

We also tested the Laplace likelihood formulation (Section~\ref{sec:prior_loss_derivation}).

\begin{table*}
\centering
\caption{Tuning of loss weights $\lambda_{\text{KL}}$ (KL divergence) and $\gamma$ (variance regularization) with Laplace likelihood. MPJPE in cm, Sharpness in cm.}
\label{tab:kl_recon_ablation}
\resizebox{\textwidth}{!}{%
\begin{tabular}{l|ccccc|ccccc|ccccc|ccccc|ccccc}
\toprule
& \multicolumn{5}{c|}{$\lambda_{\text{KL}}{=}1$} & \multicolumn{5}{c|}{$\lambda_{\text{KL}}{=}5$} & \multicolumn{5}{c|}{$\lambda_{\text{KL}}{=}10$} & \multicolumn{5}{c|}{$\lambda_{\text{KL}}{=}15$} & \multicolumn{5}{c}{$\lambda_{\text{KL}}{=}20$} \\
$\gamma$ & 1 & 3 & 5 & 7 & 9 & 1 & 3 & 5 & 7 & 9 & 1 & 3 & 5 & 7 & 9 & 1 & 3 & 5 & 7 & 9 & 1 & 3 & 5 & 7 & 9 \\
\midrule
MPJPE$\downarrow$ & 6.85 & 6.78 & 6.81 & \textbf{6.68} & 6.69 & 6.92 & 6.78 & 6.70 & 6.80 & 7.15 & 7.57 & 7.35 & 7.49 & 7.37 & 7.40 & 7.53 & 6.84 & 6.87 & 6.87 & 6.79 & 11.05 & 10.98 & 10.74 & --- & --- \\
ECE$\downarrow$ & .148 & .165 & .180 & .188 & .187 & .148 & .175 & .183 & .190 & .182 & .146 & .163 & .182 & .176 & .183 & .138 & .175 & .182 & .191 & .200 & \textbf{.124} & .155 & .164 & --- & --- \\
Sharpness & 0.94 & 0.47 & 0.31 & 0.27 & 0.24 & 0.89 & 0.43 & 0.32 & 0.28 & 0.31 & 1.23 & 0.62 & 0.45 & 0.37 & 0.32 & 1.32 & 0.49 & 0.34 & 0.29 & 0.24 & 1.83 & 0.87 & 0.68 & --- & --- \\
\bottomrule
\end{tabular}%
}
\\[2pt]
{\footnotesize Best values in \textbf{bold}. Evaluated on MM-Fi cross-subject validation split.}
\end{table*}

In summary, we therefore chose the Gaussian likelihood formulation for the prior because of our empirical results.
\section{Detailed Computational Analysis}
\label{sec:appendix_computational}

We design our model for efficient inference without sacrificing the benefits of probabilistic uncertainty quantification. To ensure a fair comparison, we implement optimized inference pipelines for all evaluated models under their best-performing configurations. All models are evaluated in a causal, online setting with batch size 1, reflecting a realistic deployment scenario.

For temporally-aware models such as MH-NFPG and mmDiff, we employ a circular buffer that maintains the last $n_t$ backbone predictions as a sliding window of temporal context. This allows the model to perform a single forward pass per frame, using cached backbone predictions until the buffer is filled and continuously updated thereafter.

We exploit batched computation for parallelism wherever possible and present the resulting inference procedures in Algorithms~\ref{alg:nfpg},~\ref{alg:diffpose}, and~\ref{alg:mmdiff}. For the normalizing flow model, we draw a batch of $N$ samples from the Laplace base distribution and propagate them through the flow in parallel, producing the full output distribution in a single forward pass. For the diffusion baselines, we parallelize hypothesis sampling across the batch dimension, while computing the conditioning only once (Table~\ref{tab:model_breakdown}). For MG-Prior + DiffPose~\cite{holmquist2023diffpose}, we use 25 denoising steps, matching the original paper's configuration. For mmDiff~\cite{fan2024diffusion}, we keep the authors' model, training schedule and sampler but use our input windows and the same temporal context of six past poses as our model. Its full diffusion schedule comprises 51 timesteps, but because the backbone already provides a partially denoised pose estimate, the first 26 steps are skipped at inference. The deterministic DDIM variant ($\eta{=}0$) further reduces this to 3 steps as in the original implementation, while the probabilistic DDPM variant ($\eta{=}1$) uses the remaining 25 steps. All models are benchmarked on an NVIDIA RTX 3090 Ti GPU.

\begin{table}[tbp]
\centering
\caption{Component-wise breakdown of model architectures (200 hypotheses).}
\label{tab:model_breakdown}
\resizebox{\columnwidth}{!}{%
\begin{tabular}{lccc}
\toprule
\textbf{Component} & \textbf{Parameters} & \textbf{GFLOPs} & \textbf{Time (ms)} \\
\midrule
\multicolumn{4}{l}{\textbf{MG-Prior}} \\
\midrule
Spatiotemporal Transformer Backbone & 8,945,152 & 27.720 & 4.988 $\pm$ 0.011 \\
Gaussian Head - Encoder & 1,575,425 & 0.003 & 0.160 $\pm$ 0.002 \\
Gaussian Head - Latent Sampling & --- & --- & 0.095 $\pm$ 0.004 \\
Gaussian Head - Decoder & 88,273 & 0.165 & 0.851 $\pm$ 0.005 \\
Aggregation & --- & --- & 0.141 $\pm$ 0.003 \\
\textbf{Total} & \textbf{10,608,850} & \textbf{27.889} & \textbf{6.235 $\pm$ 0.012} \\
\midrule
\multicolumn{4}{l}{\textbf{MG-Prior + NFPG}} \\
\midrule
Spatiotemporal Transformer Backbone & 8,945,152 & 27.720 & 4.752 $\pm$ 0.014 \\
Gaussian Head - Encoder & 1,575,425 & 0.003 & 0.147 $\pm$ 0.002 \\
Gaussian Head - Latent Sampling & --- & --- & 0.089 $\pm$ 0.002 \\
Gaussian Head - Decoder & 88,273 & 0.423 & 1.061 $\pm$ 0.010 \\
Aggregation & --- & --- & 0.140 $\pm$ 0.003 \\
Flow Conditioning Module & 7,496,800 & 0.010 & 2.367 $\pm$ 0.023 \\
Standard Laplace Sampling & --- & 0.000 & 0.139 $\pm$ 0.003 \\
Normalizing Flow & 1,533,152 & 0.611 & 2.090 $\pm$ 0.021 \\
\textbf{Total} & \textbf{19,638,802} & \textbf{28.767} & \textbf{10.786 $\pm$ 0.054} \\
\midrule
\multicolumn{4}{l}{\textbf{MG-Prior + DiffPose}} \\
\midrule
Spatiotemporal Transformer Backbone & 8,945,152 & 27.720 & 4.735 $\pm$ 0.026 \\
Gaussian Head - Encoder & 1,575,425 & 0.003 & 0.151 $\pm$ 0.003 \\
Gaussian Head - Latent Sampling & --- & --- & 0.091 $\pm$ 0.003 \\
Gaussian Head - Decoder & 88,273 & 0.423 & 1.076 $\pm$ 0.009 \\
Aggregation & --- & --- & 0.141 $\pm$ 0.003 \\
Conditioning Embedding & 13,605,760 & 16.122 & 6.252 $\pm$ 0.018 \\
Diffusion Denoiser & 2,009,678 & 20.066 & 12.654 $\pm$ 0.093 \\
\textbf{Total} & \textbf{26,224,288} & \textbf{64.335} & \textbf{25.099 $\pm$ 0.123} \\
\midrule
\multicolumn{4}{l}{\textbf{mmDiff $\eta$=0 (deterministic)}} \\
\midrule
PointTransformer Backbone & 8,009,187 & 62.806 & 12.627 $\pm$ 0.050 \\
Global Conditioning & 9,312 & 0.001 & 0.331 $\pm$ 0.004 \\
Temporal Conditioning & 37,386,048 & 0.001 & 2.799 $\pm$ 0.011 \\
GCNDiff Denoising incl. Limb Head & 1,776,326 & 0.135 & 25.202 $\pm$ 0.219 \\
\textbf{Total} & \textbf{47,180,873} & \textbf{62.942} & \textbf{40.959 $\pm$ 0.235} \\
\midrule
\multicolumn{4}{l}{\textbf{mmDiff $\eta$=1}} \\
\midrule
PointTransformer Backbone & 8,009,187 & 62.806 & 12.627 $\pm$ 0.050 \\
Global Conditioning & 9,312 & 0.001 & 0.331 $\pm$ 0.004 \\
Temporal Conditioning & 37,386,048 & 0.001 & 2.799 $\pm$ 0.011 \\
GCNDiff Denoising incl. Limb Head & 1,776,326 & 219.007 & 220.845 $\pm$ 2.517 \\
\textbf{Total} & \textbf{47,180,873} & \textbf{281.814} & \textbf{236.602 $\pm$ 2.528} \\
\bottomrule
\end{tabular}}
\end{table}

\begin{algorithm*}[tbp]
\caption{MH-NFPG Inference (Single Forward Pass)}
\label{alg:nfpg}
\small
\begin{algorithmic}[1]
\Require Point cloud sequence $\mathbf{X}$, temporal buffer $\mathcal{B}$, number of hypotheses $N$, prior samples $N_p{=}512$
\Ensure $N$ pose hypotheses $\{\hat{\mathbf{y}}^{(s)}\}_{s=1}^{N}$
\State $\mathbf{F} \gets \text{SpatiotemporalTransformer}(\mathbf{X})$ \Comment{Backbone: single forward pass}
\State $\boldsymbol{\mu}_l, \log\boldsymbol{\sigma}_l^2 \gets \text{Encoder}(\mathbf{F})$ \Comment{Latent parameters}
\State $\{\mathbf{z}_l^{(s)}\}_{s=1}^{N_p} \gets \boldsymbol{\mu}_l + \boldsymbol{\sigma}_l \odot \boldsymbol{\epsilon}, \quad \boldsymbol{\epsilon} \sim \mathcal{N}(\mathbf{0}, \mathbf{I})$ \Comment{Batched latent sampling}
\State $\{\hat{\mathbf{y}}_{\text{prior}}^{(s)}\}_{s=1}^{N_p} \gets \text{GCN}_{\text{dec}}(\{\mathbf{z}_l^{(s)}\}_{s=1}^{N_p})$ \Comment{Batched decoding}
\State $\mathbf{c} \gets \text{CondGCN}(\mathbf{F},\; \mathcal{B})$ \Comment{Features + temporal buffer $\to$ conditioning}
\State $\{\mathbf{z}^{(s)}\}_{s=1}^{N} \sim \text{Laplace}(\mathbf{0},\; 1/\sqrt{2})$ \Comment{Batched base distribution sampling}
\State $\{\hat{\mathbf{y}}^{(s)}\}_{s=1}^{N} \gets f^{-1}(\{\mathbf{z}^{(s)}\}_{s=1}^{N};\; \mathbf{c})$ \Comment{Batched inverse flow (Real NVP)}
\State $\mathcal{B}.\text{update}(\bar{\mathbf{y}}_{\text{prior}})$ \Comment{Update buffer with the prior mean (used directly)}
\State \Return $\{\hat{\mathbf{y}}^{(s)}\}_{s=1}^{N}$
\end{algorithmic}
\end{algorithm*}

\begin{algorithm*}[tbp]
\caption{MG-Prior + DiffPose Inference ($M$ Sequential Steps)}
\label{alg:diffpose}
\small
\begin{algorithmic}[1]
\Require Point cloud $\mathbf{X}$, hypotheses $N$, prior samples $N_p{=}512$, conditioning samples $S{=}32$, denoising steps $M{=}25$
\Ensure $N$ pose hypotheses $\{\hat{\mathbf{y}}^{(s)}\}_{s=1}^{N}$
\State $\mathbf{F} \gets \text{SpatiotemporalTransformer}(\mathbf{X})$ \Comment{Backbone (shared with MH-NFPG)}
\State $\boldsymbol{\mu}_l, \log\boldsymbol{\sigma}_l^2 \gets \text{Encoder}(\mathbf{F})$ \Comment{Latent parameters (shared with MH-NFPG)}
\State $\{\mathbf{z}_l^{(j)}\}_{j=1}^{N_p} \gets \boldsymbol{\mu}_l + \boldsymbol{\sigma}_l \odot \boldsymbol{\epsilon}, \quad \boldsymbol{\epsilon} \sim \mathcal{N}(\mathbf{0}, \mathbf{I})$ \Comment{Latent samples (as in MH-NFPG)}
\State $\{\hat{\mathbf{y}}_{\text{prior}}^{(j)}\}_{j=1}^{N_p} \gets \text{GCN}_{\text{dec}}(\{\mathbf{z}_l^{(j)}\}_{j=1}^{N_p})$ \Comment{Decoded prior samples}
\State $\boldsymbol{\mu}, \boldsymbol{\Sigma} \gets$ mean and covariance of $\{\hat{\mathbf{y}}_{\text{prior}}^{(j)}\}_{j=1}^{N_p}$ \Comment{Parametric MG prior}
\For{each hypothesis $s \in \{1,\ldots,N\}$} \Comment{Batched}
    \State $\{\hat{\mathbf{y}}_i\}_{i=1}^{S-1} \sim \mathcal{N}(\boldsymbol{\mu}, \boldsymbol{\Sigma}), \quad \hat{\mathbf{y}}_S \gets \boldsymbol{\mu}$ \Comment{$S$ poses including the mean}
    \State $\mathbf{c}_{s} \gets \text{Transformer}(\{\hat{\mathbf{y}}_i\}_{i=1}^{S},\; \{\log\mathcal{N}(\hat{\mathbf{y}}_i; \boldsymbol{\mu}, \boldsymbol{\Sigma})\}_{i=1}^{S})$ \Comment{Likelihood-weighted conditioning}
\EndFor
\State $\{\mathbf{x}_M^{(s)}\}_{s=1}^{N} \sim \mathcal{N}(\mathbf{0}, \mathbf{I})$ \Comment{Initialize $N$ noise samples}
\For{$t = M, \ldots, 1$} \Comment{\textbf{Sequential} denoising loop}
    \State $\boldsymbol{\epsilon}_\theta \gets \text{Denoiser}(\{\mathbf{x}_t^{(s)}\}_{s=1}^{N},\; t,\; \{\mathbf{c}_{s}\}_{s=1}^{N})$ \Comment{Batched over $N$}
    \State $\{\mathbf{x}_{t-1}^{(s)}\} \gets \text{DDPM}(\{\mathbf{x}_t^{(s)}\},\; \boldsymbol{\epsilon}_\theta,\; t)$
\EndFor
\State \Return $\{\mathbf{x}_0^{(s)}\}_{s=1}^{N}$
\end{algorithmic}
\end{algorithm*}

\begin{algorithm*}[tbp]
\caption{mmDiff Inference ($M$ Sequential Steps)}
\label{alg:mmdiff}
\small
\begin{algorithmic}[1]
\Require Point cloud $\mathbf{X}$, temporal buffer $\mathcal{B}$, hypotheses $N$, steps $M$, stochasticity $\eta$
\Ensure $N$ pose hypotheses $\{\hat{\mathbf{y}}^{(s)}\}_{s=1}^{N}$
\State $\hat{\mathbf{y}}_{\text{coarse}},\, \mathbf{F}_{\text{joint}} \gets \text{PointTransformer}(\mathbf{X})$ \Comment{Backbone + coarse pose}
\State $\mathbf{c} \gets \text{GlobalCond}(\mathbf{F}_{\text{joint}}) + \text{TemporalCond}(\mathcal{B})$ \Comment{Deterministic conditioning}
\State $\mathbf{e}_{\text{limb}} \gets \text{LimbEmbed}(\mathbf{F}_{\text{joint}})$ \Comment{Predicted limb lengths}
\State Expand $\mathbf{c},\, \mathbf{e}_{\text{limb}}$ to $N$ hypotheses \Comment{Shared across all hypotheses}
\State $\mathbf{x}_M^{(s)} \gets \hat{\mathbf{y}}_{\text{coarse}}$ for all $s$ \Comment{Start from the coarse pose}
\For{$t = M, \ldots, 1$} \Comment{\textbf{Sequential} denoising loop}
    \State $\boldsymbol{\epsilon}_\theta \gets \text{Denoiser}(\{\mathbf{x}_t^{(s)}\}_{s=1}^{N},\; t,\; \mathbf{c},\; \mathbf{e}_{\text{limb}})$ \Comment{Batched over $N$}
    \State $\{\mathbf{x}_{t-1}^{(s)}\} \gets \text{DDIM/DDPM}(\{\mathbf{x}_t^{(s)}\},\; \boldsymbol{\epsilon}_\theta,\; t,\; \eta)$ \Comment{$\eta{=}0$: det., $\eta{=}1$: stoch.}
\EndFor
\State $\mathcal{B}.\text{update}(\hat{\mathbf{y}}_{\text{coarse}})$ \Comment{Update temporal buffer}
\State \Return $\{\mathbf{x}_0^{(s)}\}_{s=1}^{N}$
\end{algorithmic}
\end{algorithm*}

In Algorithm~\ref{alg:nfpg}, all operations on $N$ hypotheses are batched along a single tensor dimension and executed in parallel, with the conditioning computed once and shared across all hypotheses. MH-NFPG uses the estimates of the $N_p{=}512$ decoded prior samples directly, and their mean fills the temporal buffer. In contrast, both diffusion baselines require $M$ sequential denoising steps (25, or 3 for mmDiff with $\eta{=}0$). The two diffusion models differ in their conditioning. For MG-Prior + DiffPose (Algorithm~\ref{alg:diffpose}), we stay as close to the original as our setting allows. The original draws joint candidates from the 2D heatmaps of an RGB keypoint detector and conditions on them together with the heatmap maximum. Radar point clouds provide no such heatmaps, so we instead form the parametric MG-Prior distribution $\mathcal{N}(\boldsymbol{\mu}, \boldsymbol{\Sigma})$ from the same $N_p{=}512$ samples, at the same fidelity as for MH-NFPG, and condition each hypothesis on its mean and $S{-}1$ poses drawn from it, weighted by their likelihood, with $S{=}32$ as in the original configuration. As in the original implementation, the conditioning vector of each hypothesis is computed once and kept fixed across the denoising steps, so the conditioning is batched across all $N$ hypotheses while the denoising loop remains sequential. mmDiff (Algorithm~\ref{alg:mmdiff}) instead uses deterministic conditioning computed once from per-joint backbone features and temporal context, shared across all hypotheses. Here, all hypotheses start from the coarse pose, and stochasticity arises only from the noise injected at every step ($\eta{=}1$).

\section{Qualitative Results}

\begin{figure*}[t]
    \centering
    \includegraphics[width=\textwidth,height=0.85\textheight,keepaspectratio]{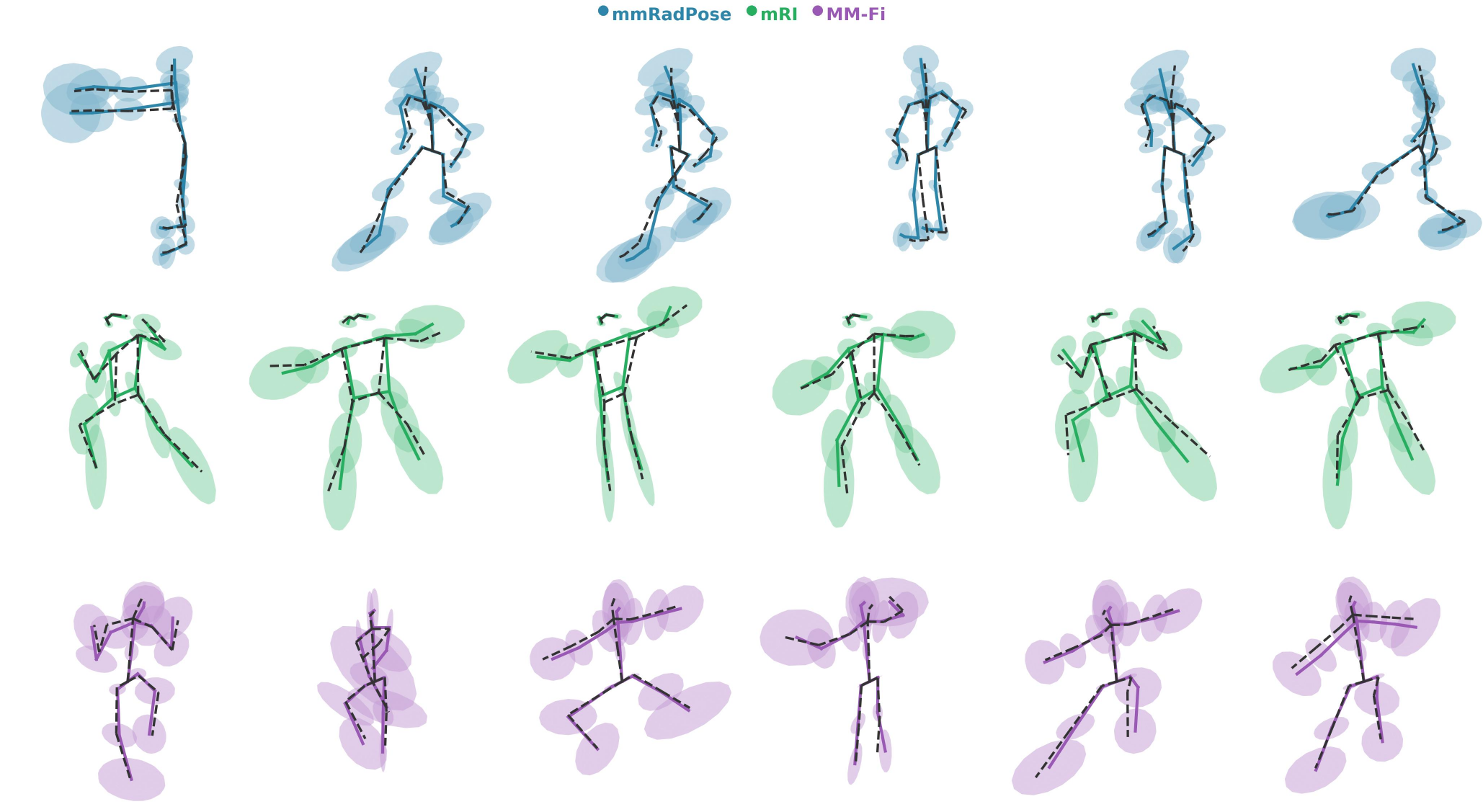}
    \caption{Qualitative results of MH-NFPG across mmRadPose, mRI, and MM-Fi. Predicted poses with per-joint 90\% confidence ellipsoids consistently contain the ground truth, confirming calibrated uncertainty across diverse poses and radar configurations.}
    \label{fig:poses_datasets}
\end{figure*}

\begin{figure*}[t]
    \centering
    \includegraphics[width=\textwidth,height=0.83\textheight,keepaspectratio]{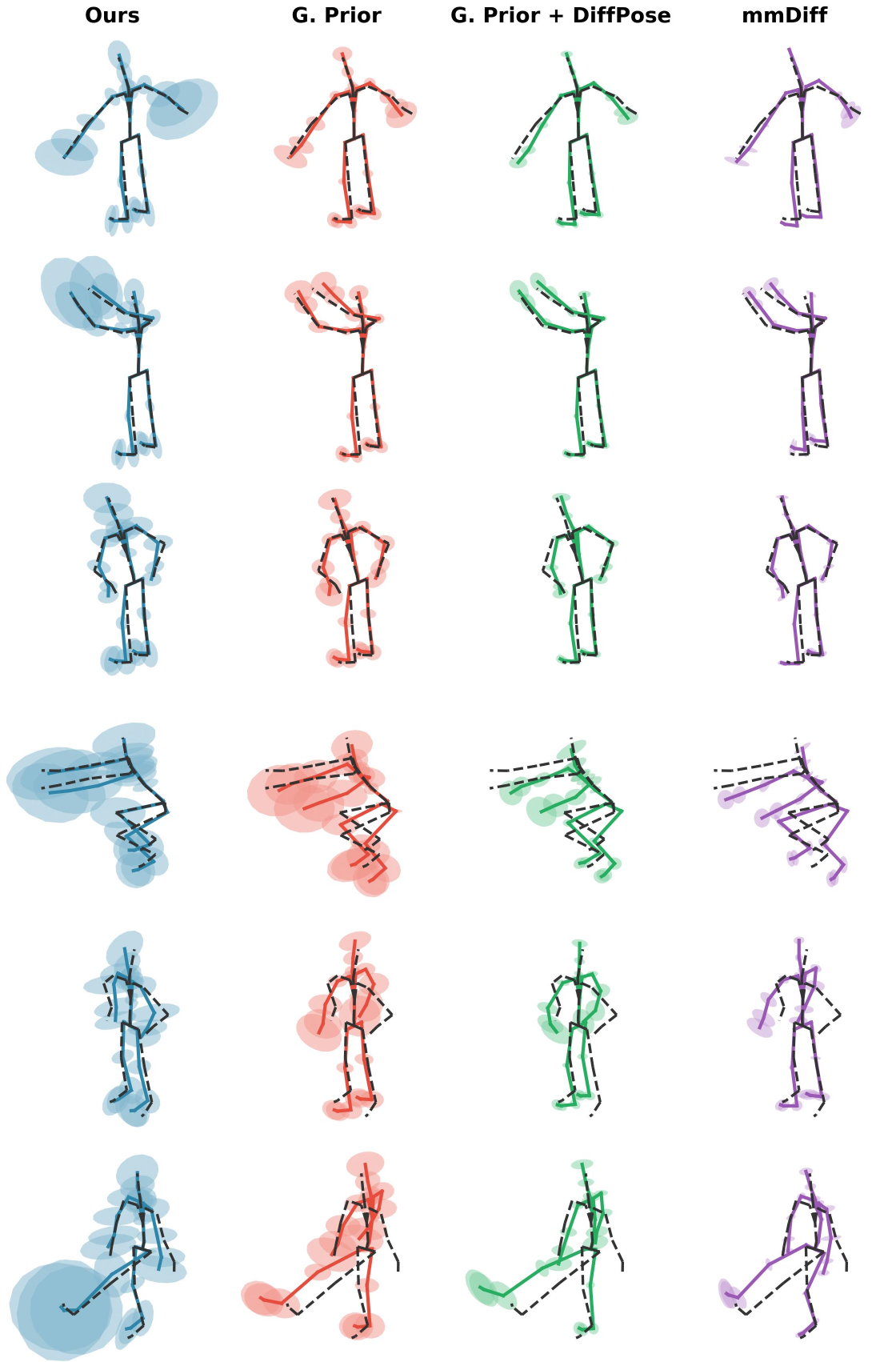}
    \caption{Extended qualitative comparison on mmRadPose. Predictions (solid, colored) and ground truth (dashed, black) with 90\% confidence ellipsoids for MH-NFPG, MG-Prior, MG-Prior + DiffPose, and mmDiff. The first three rows show easy examples, while the last three rows show ambiguous examples with occlusions and more complex multipath effects. MH-NFPG consistently produces well-calibrated confidence regions.}
    \label{fig:qualitative_appendix}
\end{figure*}

Figure~\ref{fig:poses_datasets} shows MH-NFPG predictions across all three datasets, demonstrating that calibrated uncertainty estimates generalize across diverse poses, subjects, and radar configurations, with ground-truth joints consistently falling within the predicted confidence ellipsoids. Figure~\ref{fig:qualitative_appendix} provides an extended comparison against DiffPose and mmDiff on mmRadPose. The first three rows show easy examples, while the last three rows show ambiguous examples with occlusions and more complex multipath effects. MH-NFPG produces uncertainty estimates appropriate for the ambiguity of each sample and recognizes which joints have smaller radar cross-sections and are therefore subject to noisier observations, assigning them larger uncertainties accordingly.

\section{Additional Uncertainty Analysis}

\begin{figure}[t]
    \centering
    \includegraphics[width=0.7\linewidth]{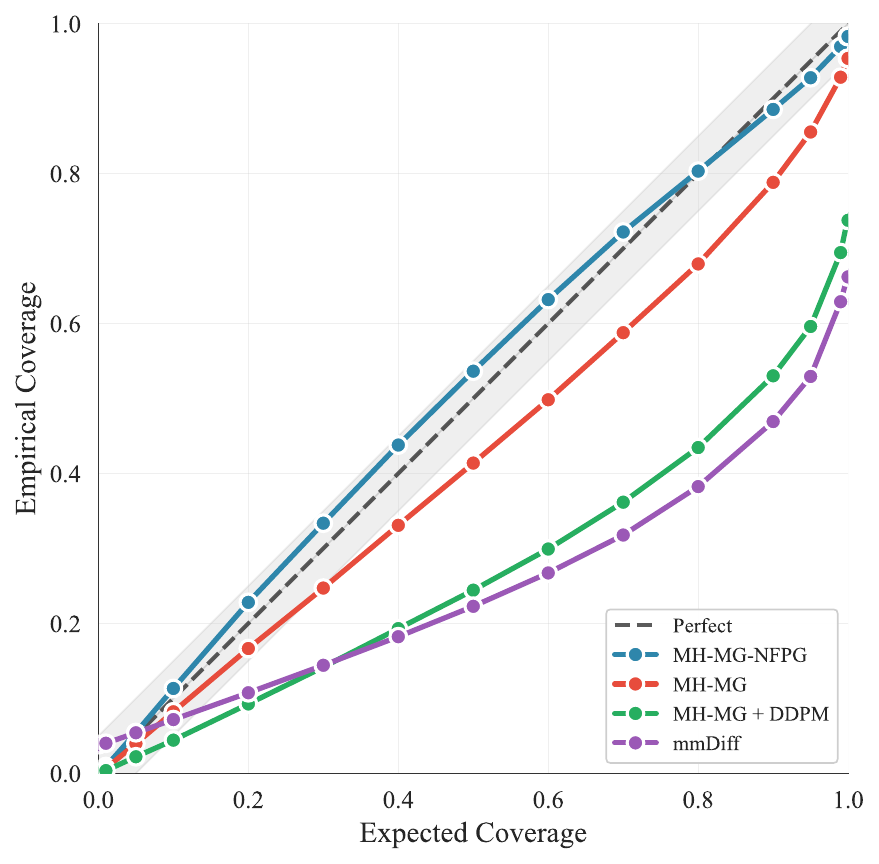}
    \caption{Reliability diagram on mmRadPose. MH-NFPG (blue) closely follows the perfect calibration diagonal (dashed), while diffusion baselines exhibit severe undercoverage.}
    \label{fig:calibration_curve}
\end{figure}

Figure~\ref{fig:calibration_curve} shows the reliability diagram on mmRadPose, plotting empirical coverage against expected coverage for all evaluated probabilistic models. A perfectly calibrated model follows the diagonal. MH-NFPG closely tracks the diagonal across all confidence levels, confirming well-calibrated uncertainty estimates. In contrast, MG-Prior + DiffPose and mmDiff ($\eta{=}1$) fall far below the diagonal, indicating severe overconfidence where predicted confidence intervals consistently fail to cover the ground truth.

Tables~\ref{tab:joint_var_mmradpose} and~\ref{tab:joint_var_mri} report the per-joint sharpness for MH-NFPG across all three datasets, computed from 200 hypotheses averaged over all test frames. Across all datasets, a consistent pattern emerges. Joints with smaller radar cross-sections (hands, feet) exhibit higher per-joint sharpness values, while joints with larger reflective surface area show lower uncertainty. This aligns with physical expectations, as smaller radar cross-sections yield weaker and less consistent reflections and are more susceptible to self-occlusion and multipath effects. Crucially, MH-NFPG dynamically expresses this variation in its per-joint uncertainties, capturing meaningful physical structure of radar observability rather than arbitrary noise, as additionally visualized by the confidence ellipsoids in Figures~\ref{fig:qualitative_actions} and~\ref{fig:qualitative_comparison} of the main paper. Furthermore, the overall per-joint sharpness is notably lower on mmRadPose, which uses clean optical motion capture ground truth, compared to mRI and MM-Fi, where ground truth is obtained from a vision-based pose estimator. This indicates that the model additionally adapts its uncertainty to reflect the precision of the supervision signal.

\begin{table}
\centering
\caption{Per-joint sharpness -- mmRadPose (31\,835 test frames).}
\label{tab:joint_var_mmradpose}
\begin{tabular}{l r}
\toprule
\textbf{Joint} & \textbf{Sharpness (cm)} \\
\midrule
RightUpLeg     &  0.699 \\
RightLeg       &  3.067 \\
RightFoot      &  4.842 \\
RightToeBase   &  5.050 \\
RightToeEnd    &  5.243 \\
LeftUpLeg      &  0.657 \\
LeftLeg        &  3.190 \\
LeftFoot       &  4.930 \\
LeftToeBase    &  5.691 \\
LeftToeEnd     &  5.368 \\
Spine          &  0.409 \\
Spine1         &  2.045 \\
RightShoulder  &  3.656 \\
RightArm       &  3.793 \\
RightForeArm   &  4.468 \\
RightHand      &  5.975 \\
RightHandEnd   &  7.386 \\
LeftShoulder   &  3.447 \\
LeftArm        &  3.889 \\
LeftForeArm    &  4.492 \\
LeftHand       &  6.848 \\
LeftHandEnd    &  9.409 \\
Neck           &  4.043 \\
Head           &  4.574 \\
HeadEnd        &  6.166 \\
\bottomrule
\end{tabular}
\end{table}

\begin{table}
\centering
\caption{Per-joint sharpness -- mRI (26\,511 test frames) and MM-Fi (62\,424 test frames).}
\label{tab:joint_var_mri}
\resizebox{\columnwidth}{!}{%
\begin{tabular}{l r @{\hskip 2em} l r}
\toprule
\multicolumn{2}{c}{\textbf{mRI}} & \multicolumn{2}{c}{\textbf{MM-Fi}} \\
\textbf{Joint} & \textbf{Sharpness (cm)} & \textbf{Joint} & \textbf{Sharpness (cm)} \\
\midrule
L\_Eye         &  1.006 & R\_Hip         &  1.661 \\
R\_Eye         &  1.058 & R\_Knee        &  3.771 \\
L\_Ear         &  2.562 & R\_Ankle       &  6.283 \\
R\_Ear         &  2.329 & L\_Hip         &  1.600 \\
L\_Shoulder    &  3.596 & L\_Knee        &  3.694 \\
R\_Shoulder    &  3.604 & L\_Ankle       &  5.795 \\
L\_Elbow       &  5.331 & Spine          &  2.840 \\
R\_Elbow       &  5.335 & Thorax         &  5.679 \\
L\_Wrist       &  8.019 & Neck           &  6.645 \\
R\_Wrist       &  8.050 & Head           &  6.836 \\
L\_Hip         &  5.204 & L\_Shoulder    &  5.907 \\
R\_Hip         &  5.173 & L\_Elbow       &  6.775 \\
L\_Knee        &  8.548 & L\_Wrist       &  8.960 \\
R\_Knee        &  8.121 & R\_Shoulder    &  5.898 \\
L\_Ankle       & 10.493 & R\_Elbow       &  6.585 \\
R\_Ankle       & 10.837 & R\_Wrist       &  8.343 \\
\bottomrule
\end{tabular}%
}
\end{table}

\clearpage

\end{document}